\documentclass[10pt]{article}
\usepackage[preprint]{tmlr}
\usepackage{booktabs}
\usepackage{graphicx}
\usepackage{multirow}
\usepackage{amsmath}
\usepackage{amssymb}
\usepackage{algorithm}
\usepackage{algpseudocode}
\usepackage{float}
\usepackage{tikz}
\usepackage{url}
\usepackage{hyperref}
\usepackage[table]{xcolor}
\usepackage{xspace}

\usetikzlibrary{arrows.meta}
\definecolor{rsupports}{HTML}{1A1A1A}
\definecolor{rrefines}{HTML}{8B4513}
\definecolor{rsequence}{HTML}{2C5F5F}
\newcommand{\typename}[1]{\texttt{#1}}

\definecolor{darkblue}{rgb}{0, 0, 0.5}
\hypersetup{pdfauthor={Dan C. Hsu and Luke Lu}, pdftitle={Scoped Verification for Reliable Long-Horizon Agentic Context Evolution under Distribution Shift}, colorlinks=true, citecolor=darkblue, linkcolor=darkblue, urlcolor=darkblue}

\def\month{MM}
\def\year{YYYY}
\def\openreview{\url{https://openreview.net/forum?id=XXXX}}

\title{Scoped Verification for Reliable Long-Horizon Agentic Context Evolution under Distribution Shift}

\author{\name Dan C. Hsu \email dan@redmindresearch.org \\
      \addr RedMind Research Foundation; National Taiwan University
      \AND
      \name Luke Lu\thanks{Corresponding author.} \email luke@redmindresearch.org \\
      \addr RedMind Research Foundation}

\begin{document}

\maketitle

\begin{abstract}
Deployed LLM agents rely on agentic context, the model-external textual control content assembled by an operational harness. In this work, the mutable component of that context is a persistent system-level instruction that is updated from operational experience while the model, tools, and harness remain fixed. Over long evolution horizons, flat-text maintenance makes verification increasingly difficult as accumulated instructions grow and interact. We propose Graph-Regularized Agentic Context Evolution (GRACE), which maintains the persistent instruction component as a typed semantic graph and validates proposed updates within the local typed neighborhoods of modified nodes. Accepted graph updates are reconstructed as incremental edits to the textual instruction checkpoint used at deployment. We evaluate GRACE within a fixed telecom agent harness derived from $\tau^2$-bench under a controlled distribution-shift protocol. Across five independent replications, GRACE improves strict reliability, measured by pass\textasciicircum{}3, from the Gemini 2.5 Flash zero-shot value of 0.091 to 0.661$\pm$0.158 at the final checkpoint. This exceeds a Gemini 3.1 Pro zero-shot reference of 0.242 on the same held-out set, while the flat-text HCE baseline finishes at 0.191$\pm$0.051. Replication-level tests separate GRACE from the flat-text baseline at every reported checkpoint, and the flat-text instruction is 1.6$\times$ longer than GRACE's at deployment, so GRACE's extra offline verification cost is recovered after roughly 1{,}200 agent turns, within the first one or two experience batches of deployment. The evidence is confined to one domain, one agent model, and one harness, and the evolution conditions are not matched in internal LLM-call budget; within that scope, the results identify two requirements for reliable long-horizon context evolution, a structural substrate that makes verification local and a consolidation mechanism that keeps accumulated instruction content usable. Code and reproduction runner: \url{https://github.com/RedMind-Research/GRACE}.
\end{abstract}

\section{Introduction}\label{sec:intro}
As large language models have become better at following natural-language instructions, model-external text has become a primary control surface for deployed agents \citep{khattab2024dspy, zhang2026ace}.\footnote{An LLM assistant was used to polish author-drafted text and to help write and check the analysis scripts; see Appendix~\ref{app:llm-use}.} In operational agent systems, this text is assembled with task inputs, tool observations, and harness-provided information before each model call \citep{yao2024taubench, barres2026tau2bench}. We refer to the resulting inference-time input as the agent context. In the setting studied here, the mutable component of this context is a persistent system-level instruction that specifies the agent's role, behavioral constraints, procedural guidance, and domain assumptions. When task distributions shift, this persistent instruction can encode stale or incomplete assumptions about the environment it is meant to govern. We study context evolution as the process of revising this persistent context component from operational experience while the model, tools, and harness remain fixed. The evolving artifact may be a playbook \citep{zhang2026ace}, an adaptive memory \citep{suzgun2026dc}, or a set of structured guidelines \citep{pei2025scope}. Across these forms, prior work has shown that iterative updates can improve agent behavior \citep{zhang2026ace, shinn2023reflexion}.

However, over extended evolution horizons, previously accumulated improvements can be undermined by inconsistencies introduced in subsequent steps. Because the persistent instruction is included in the agent context throughout deployment, degradation in this artifact can compromise operational reliability as well as update efficiency. Context collapse during full-document rewriting has been documented as one failure mode in agentic context engineering \citep{zhang2026ace}, alongside memory truncation in adaptive memory \citep{suzgun2026dc} and guideline conflict accumulation in prompt evolution \citep{pei2025scope}. These failures point to a shared structural limitation. Flat-text maintenance leaves relationships among rules, constraints, and procedures implicit in the document's linear order, which makes verification increasingly dependent on long-context processing as the artifact grows \citep{liu2024lost}. Prior work has shown that model-external guidance can be iteratively improved, but it has not directly isolated whether verification remains effective when the persistent instruction grows over long evolution horizons.

\begin{table}[!t]
\centering
\scriptsize
\setlength{\tabcolsep}{3pt}
\renewcommand{\arraystretch}{1.08}
\begin{tabular}{p{3.15cm} p{1.9cm} p{2.15cm} p{2.15cm} p{1.95cm} p{2.15cm}}
\toprule
\textbf{Method} & \textbf{Evolving artifact} & \textbf{Explicit check on the artifact} & \textbf{Where the check is applied} & \textbf{Consolidation} & \textbf{Update horizon evaluated} \\
\midrule
Reflexion \citep{shinn2023reflexion} & list of self-reflections & none & -- & FIFO truncation (1--3 items) & a few trials of one task \\
DSPy / MIPRO \citep{khattab2024dspy} & per-module prompts & none (score-gated selection) & -- & none & one offline compile \\
GEPA \citep{agrawal2026gepa} & per-module prompts & none (score-gated selection) & -- & implicit in whole-prompt rewrite & offline search budget \\
ExpeL \citep{zhao2024expel} & list of insights & none & -- & vote-based pruning & offline train-then-test \\
Agent Workflow Memory \citep{wang2025awm} & list of workflows & soft ``avoid overlap'' instruction & -- & none & online growth, not measured \\
MemoryBank \citep{zhong2024memorybank} & dialogue log + summaries & none & -- & hierarchical summaries, forgetting & single evaluation point \\
Mem0 \citep{chhikara2025mem0} & memory facts; graph variant & dedup and conflict resolution inside the update LLM & incoming item vs.\ retrieved candidates & update / delete; invalidate relations & single QA snapshot \\
A-MEM \citep{xu2025amem} & linked notes & none & edits limited to embedding neighbors of a new note & none & single QA snapshot \\
Zep \citep{rasmussen2025zep} & bi-temporal knowledge graph & entity dedup, temporal invalidation & incoming episode vs.\ retrieved candidates & merge, summaries & single QA snapshot \\
ACE \citep{zhang2026ace} & itemized playbook & embedding-based dedup & whole playbook & dedup only & multi-round; collapse studied \\
\midrule
GRACE (this paper) & persistent system instruction & contradiction and redundancy detection & $k$-hop typed neighborhood of the edit & merge / modify repairs & 10 updates under distribution shift \\
\bottomrule
\end{tabular}
\caption{\textbf{How related systems maintain an evolving model-external artifact.} Cells summarize each paper's own description of its mechanism; ``none'' means the paper describes no such step, not that the system could not support one. Retrieved memories are read selectively at inference, whereas the instruction GRACE maintains is read in full on every agent turn.}
\label{tab:related-compare}
\end{table}

We propose Graph-Regularized Agentic Context Evolution (GRACE), which maintains a typed semantic graph as the intermediate substrate for evolving the persistent instruction component of agent context. GRACE represents atomic instruction units as graph nodes, encodes structural relationships among them as typed directed edges, and uses graph locality to validate each proposed update near the affected nodes. Accepted graph updates are reconstructed as incremental text edits to the deployed instruction checkpoint. This paper studies the representation substrate within a fixed deployment interface. The agent model, tools, harness, diagnosis procedure, and evaluation set are held fixed, while the representation and verification mechanism vary. We evaluate GRACE in the telecom customer-service domain derived from $\tau^2$-bench \citep{barres2026tau2bench}, using its operational harness with fixed tool interfaces and evaluation criteria. Under a controlled shift protocol across 10 evolution batches, the main comparison uses three conditions with the same diagnosis procedure but different representation and verification mechanisms: GRACE, GRACE without structural analysis, and HCE. Across replicated runs, GRACE remains ahead at the later checkpoints, indicating that structure-enabled verification and consolidation both matter for sustained context evolution.

In summary, the paper makes three contributions. First, GRACE introduces a graph-regularized substrate for evolving the persistent instruction component of agent context and performs scoped structural validation at each evolution step. Second, the evaluation protocol introduces alternating distribution shift, so the same trajectory tests both within-phase improvement and cross-phase retention. Third, replicated experiments in the telecom domain show that GRACE outperforms a flat-text context-evolution baseline and a graph-based ablation without structural analysis. The ablation separates two failure modes. Contradiction avoidance helps, but sustained improvement also requires consolidation of the growing instruction substrate. We state the boundaries of this evidence explicitly: it covers one domain, one agent model, and one harness; the three evolution conditions are matched in diagnosis and update schedule but not in internal LLM-call budget; and the mechanism audits are LLM-judged and correlational. Section~\ref{sec:limitations} discusses what the results do and do not establish.

\section{Related Work}\label{sec:related}

\paragraph{Context evolution and persistent control artifacts.}
Optimizing the textual artifacts that steer LLM behavior has progressed from prompt-level refinement to long-horizon context evolution. Prompt optimization and self-refinement methods such as DSPy \citep{khattab2024dspy}, GEPA \citep{agrawal2026gepa}, and Reflexion \citep{shinn2023reflexion} demonstrate effective short-horizon improvement. GEPA is the closest optimizer in spirit, since it rewrites module instructions from reflection over execution traces, but each rewrite regenerates the whole instruction and is accepted on held-out score alone; none of these methods maintains a persistent context component through repeated batch-level updates or checks the artifact itself for internal consistency, and they address a different optimization setting. More recent work treats model-external guidance as a persistent artifact that accumulates knowledge across episodes. ACE \citep{zhang2026ace} introduces incremental delta updates to prevent context collapse, SCOPE \citep{pei2025scope} applies conflict resolution and subsumption pruning on dual-stream memory, and Dynamic Cheatsheet \citep{suzgun2026dc} accumulates reusable guidance over time. We focus on a narrower question inside one fixed harness, namely whether the representation substrate used by the evolution operator can sustain verification as the persistent instruction component grows.

Broader agent-evolution frameworks address tool, workflow, or environment adaptation. A-Evolve \citep{lin2026aevolve} formalizes cross-episode updates over a persistent artifact state. Our work shares this cross-episode perspective, but focuses specifically on the persistent system-level instruction inside agent context and on controlled instantiations of the same evolution interface. Across the context-evolution methods above, the evolving artifact is typically maintained as flat text, and the question of whether an alternative representation substrate can sustain verification effectiveness over extended horizons remains underexplored.

\paragraph{Relation to ACE.}
\textbf{How ACE works.} ACE \citep{zhang2026ace} is the closest prior system. It keeps the evolving context as itemized bullets, each with an identifier and counters of how often it was marked helpful or harmful. A generator produces trajectories, a reflector extracts lessons, and a curator merges the resulting bullets deterministically as compact delta entries. Its grow-and-refine step appends new bullets, updates existing ones in place, and prunes redundancy by comparing bullets through semantic embeddings.

\textbf{What GRACE shares.} Full-document rewriting is harmful, and updates should be incremental. Our HCE baseline adopts exactly this incremental-update contract on flat text, with an explicit conflict-and-duplication guard (Appendix~\ref{app:prompt-hce}).

\textbf{Where GRACE differs.} First, the substrate is a typed graph with explicit \typename{supports}, \typename{refines}, and \typename{sequence} relations rather than an ordered list, so relationships that ACE leaves implicit in item order and metadata become inspectable structure. Second, verification is scoped by graph locality around the touched nodes rather than applied to the whole playbook; this is the property we argue matters as the artifact grows. Third, structural analysis detects contradictions as well as redundancy, whereas ACE's refinement step addresses redundancy only, and GRACE repairs both through schema-admissible merge and modify operations.

\textbf{Why no head-to-head.} ACE's reflector and curator constitute its own diagnosis procedure. A faithful comparison would vary diagnosis and representation jointly. Our design holds diagnosis fixed across conditions so that differences can be attributed to representation and verification. HCE should therefore be read as an ACE-style incremental flat-text updater under our shared diagnosis, not as a reimplementation of ACE. A budget-matched comparison against the full ACE pipeline remains future work.

\paragraph{Agent memory systems.}
A related line of work maintains a growing external memory for an agent and has begun to add explicit maintenance operations. Table~\ref{tab:related-compare} summarizes the mechanisms.

\textbf{Systems with maintenance operations.} Mem0 \citep{chhikara2025mem0} stores itemized memory facts and lets the extraction LLM add, update, delete, or ignore each candidate, so deduplication and conflict handling happen implicitly inside the update step; its graph variant marks superseded relations invalid rather than removing them. Zep \citep{rasmussen2025zep} builds a bi-temporal knowledge graph with entity deduplication and temporal invalidation of contradicted facts at ingestion. MemoryBank \citep{zhong2024memorybank} consolidates a chronological log through hierarchical summaries and Ebbinghaus-style forgetting.

\textbf{Systems that only accumulate.} A-MEM \citep{xu2025amem} links Zettelkasten-style notes and lets new notes rewrite the attributes of their embedding neighbors, but describes no deduplication, conflict check, or pruning. Agent Workflow Memory \citep{wang2025awm} and ExpeL \citep{zhao2024expel} accumulate induced workflows or insights as flat lists, with at most a soft instruction against overlap or a vote-based pruning counter.

\textbf{What differs from GRACE.} These systems share GRACE's concern with keeping an accumulating artifact usable, and the local, LLM-mediated edits of Mem0, Zep, and A-MEM around newly inserted items are the closest mechanistic relatives of scoped structural analysis. Three differences matter here. The artifact is retrieved memory rather than a persistent instruction read in full on every turn. Where verification exists, it is applied at ingestion to the incoming item against retrieved candidates, not to the typed neighborhood of an edit inside a schema-constrained graph. Evaluation is a single retrieval-QA snapshot after ingestion, so degradation of the artifact across many update rounds is not measured.

\paragraph{Structured knowledge representation for LLMs.}
Graph structures have been widely adopted in LLM systems for knowledge organization, memory, and retrieval. GraphRAG \citep{edge2024graphrag} constructs entity-relation graphs to support multi-hop question answering, and HippoRAG \citep{gutierrez2024hipporag} builds knowledge graphs as long-term memory for associative retrieval. These applications demonstrate that graphs can organize knowledge that LLMs operate over, but their use as an evolving substrate for persistent instruction content remains underexplored. GRACE addresses this gap by using a typed semantic graph as the intermediate representation on which context evolution and structural validation operate.

\section{Problem Formulation}\label{sec:formulation}

We study long-horizon agentic context evolution under a fixed operational harness. Let $\pi(a \mid c)$ denote the frozen action-selection policy induced by the LLM, tools, and harness when supplied with agent context $c$. The model parameters, tool interfaces, harness logic, and context-assembly procedure are held fixed throughout the evolution trajectory.

In this work, the mutable component of the agent context is a persistent system-level instruction, denoted by $\ell_t$ after the $t$-th batch-level update. This instruction is included in the context supplied to the agent, but it is not itself the action-selection policy. At inference time, the harness assembles a task-specific context
\begin{equation}\label{eq:assemble}
c_t(h,x) = \mathrm{Assemble}(\ell_t, h, x),
\end{equation}
where $h$ denotes harness-provided information and $x$ denotes the task or user input. The full task-specific context changes across episodes through $h$ and $x$, while only $\ell_t$ is updated across evolution batches.

We model context evolution as a closed-loop cross-episode update process, following the persistent artifact update perspective of \citet{lin2026aevolve}. We use \emph{substrate} to denote the mutable representation used during offline evolution. The substrate is the internal form in which the persistent instruction is maintained, edited, and validated before the next instruction checkpoint is reconstructed. Let $s_t$ denote this substrate at update step $t$. The initial instruction $\ell_0$ is shared across all conditions, and the initial substrate $s_0$ is initialized from $\ell_0$. For text-maintained methods, $s_t$ is the instruction text itself. For GRACE, $s_t = G_t$ is a graph-regularized substrate from which $\ell_t$ is reconstructed.

For each update step $t \in \{1,\ldots,T\}$, the agent is executed on tasks drawn from the batch distribution $\mathcal{D}_t$. Execution uses the frozen action-selection policy $\pi$ under contexts assembled from the previous instruction checkpoint $\ell_{t-1}$. The resulting interaction traces are
\begin{equation}\label{eq:experience}
\mathcal{E}_t = F_{\mathrm{Experience}}\big(\pi,\ \mathrm{Assemble}(\ell_{t-1},\cdot,\cdot),\ \mathcal{D}_t\big).
\end{equation}
The diagnostic observation is then computed as
\begin{equation}\label{eq:obs}
o_t = F_{\mathrm{Diagnose}}(\mathcal{E}_t).
\end{equation}

The evolution function uses the diagnostic observation to update the substrate and produce the next instruction checkpoint:
\begin{equation}\label{eq:evolve}
(s_t, \ell_t) \leftarrow F_{\mathrm{Evolve}}(s_{t-1}, \ell_{t-1}, o_t).
\end{equation}
For a text-maintained context-evolution baseline,
\begin{equation}\label{eq:evolve-text}
\ell_t \leftarrow F^{\mathrm{text}}_{\mathrm{Evolve}}(\ell_{t-1}, o_t).
\end{equation}
For GRACE,
\begin{equation}\label{eq:evolve-graph}
(G_t, \ell_t) \leftarrow F^{\mathrm{graph}}_{\mathrm{Evolve}}(G_{t-1}, \ell_{t-1}, o_t).
\end{equation}
The graph $G_t$ is used only during offline editing and validation. Deployment uses the reconstructed textual instruction $\ell_t$, which is assembled with $h$ and $x$ to form the agent context supplied to $\pi$.

We hold $F_{\mathrm{Experience}}$ and $F_{\mathrm{Diagnose}}$ fixed across all conditions. The comparison therefore isolates how each method maintains, validates, and reconstructs the evolving instruction component through $F_{\mathrm{Evolve}}$. In this work, GRACE is the proposed graph-regularized realization of this operator, introduced in \S\ref{sec:method}. The empirical protocol in \S\ref{sec:setup} specifies the controlled shift schedule, checkpoint evaluation procedure, and reliability metrics used to evaluate the resulting deployed context checkpoints. Throughout the paper, \emph{context evolution} refers to the batch-level process that updates the persistent instruction component of the agent context. The term \emph{policy} is reserved for the frozen action-selection policy $\pi(a \mid c)$, except when discussing terminology used in prior work.

\begin{figure}[t]
\begin{center}
\vspace{-2.5em}
\includegraphics[width=\linewidth]{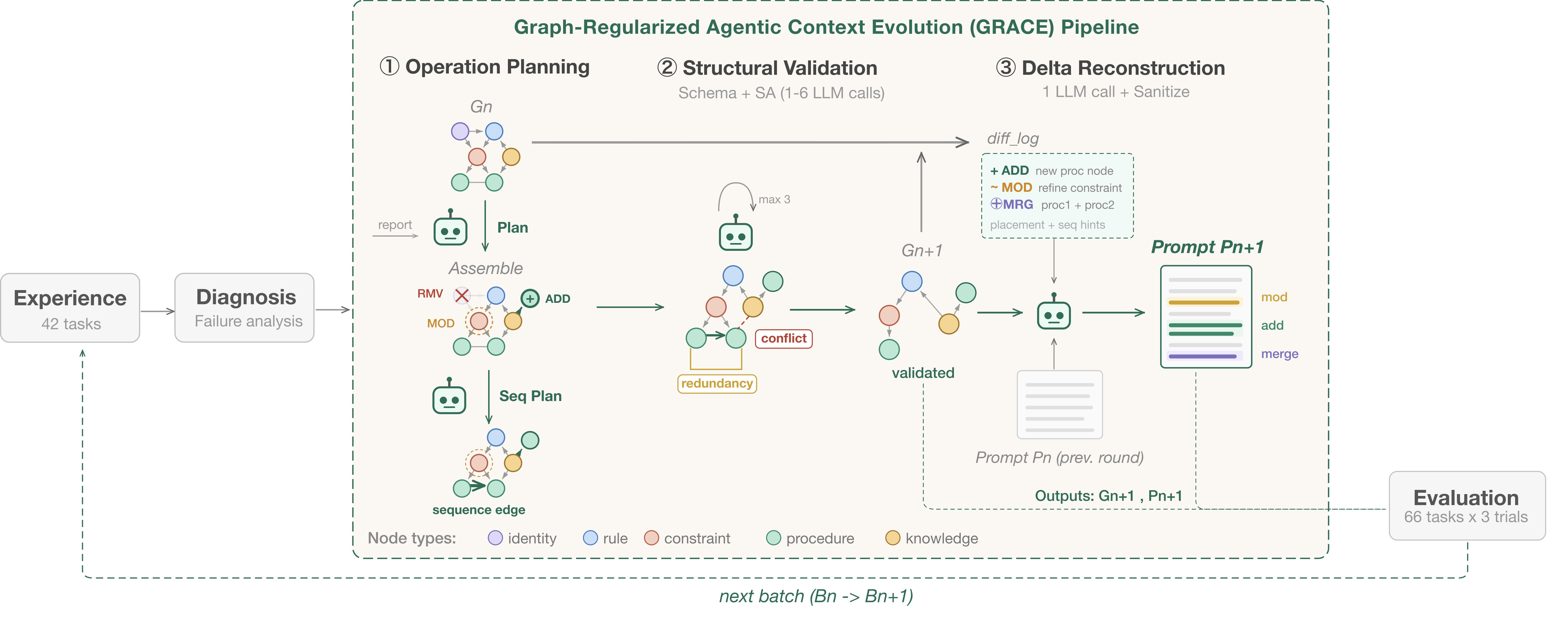}
\vspace{-2em}
\end{center}
\caption{\textbf{Overview of the GRACE evolution pipeline.} Given a diagnosis report and the current graph $G_{t-1}$, the pipeline produces an updated graph $G_t$ and corresponding instruction checkpoint $\ell_t$ through three stages: Operation Planning, Structural Validation, and Delta Reconstruction. Sequence planning is performed as a substep within Operation Planning, while Structural Validation combines deterministic schema checks with iterative structural analysis (SA). The node legend uses the informal labels \emph{rule}, \emph{constraint}, and \emph{procedure} for sub-kinds of $\mathsf{norm}$ nodes; the formal schema has the three object types of Section~\ref{sec:graph-repr}. The figure writes $P_n$, $G_n$, and $B_n$ for the instruction checkpoint $\ell_t$, the graph $G_t$, and the experience batch.}\label{fig:pipeline}
\end{figure}

\section{Method: Graph-Regularized Agentic Context Evolution}\label{sec:method}
In the notation of \S\ref{sec:formulation}, GRACE instantiates the evolution substrate $s_t$ as a typed graph $G_t$ (Figure~\ref{fig:pipeline}). Nodes encode atomic instruction units extracted from the persistent system-level instruction $\ell_t$, and edges encode structural relations among those units. The deployed artifact remains textual. After schema-constrained editing and local structural validation, GRACE reconstructs the next instruction checkpoint $\ell_t$, which is assembled into the agent context at inference time.

\subsection{Graph-Regularized Substrate}\label{sec:graph-repr}
\begin{figure}[!htbp]
    \centering
    \begin{tikzpicture}[
      >={Stealth[length=2.2mm, width=1.8mm]},
      type/.style={
        circle, draw=black, line width=1pt,
        minimum size=2cm, inner sep=0pt,
        font=\small
      },
      rel/.style={->, line width=0.8pt},
      lbl/.style={font=\footnotesize},
    ]

      %% ---- Object type nodes -----------------------------------------
      \node[type] (identity)  at (-4.5, 0) {\typename{identity}};
      \node[type] (norm)      at (0, 0)    {\typename{norm}};
      \node[type] (knowledge) at (4.5, 0)  {\typename{knowledge}};

      %% ---- Cross-type edges ------------------------------------------
      \draw[rel, color=rsupports] (identity) --
        node[lbl, above, color=rsupports] {\typename{supports}} (norm);
      \draw[rel, color=rsupports] (knowledge) --
        node[lbl, above, color=rsupports] {\typename{supports}} (norm);

      %% ---- Self-loop on identity -------------------------------------
      \draw[rel, color=rrefines] (identity)
        to[out=118, in=62, looseness=3.5]
        node[lbl, above, color=rrefines] {\typename{refines}} (identity);

      %% ---- Self-loops on norm ----------------------------------------
      \draw[rel, color=rsupports] (norm)
        to[out=118, in=62, looseness=3.5]
        node[lbl, above, color=rsupports] {\typename{supports}} (norm);
      \draw[rel, color=rrefines] (norm)
        to[out=123, in=57, looseness=5.9]
        node[lbl, above, color=rrefines] {\typename{refines}} (norm);
      \draw[rel, color=rsequence] (norm)
        to[out=128, in=52, looseness=7.5]
        node[lbl, above, color=rsequence] {\typename{sequence}} (norm);

      %% ---- Self-loops on knowledge -----------------------------------
      \draw[rel, color=rsupports] (knowledge)
        to[out=118, in=62, looseness=3.5]
        node[lbl, above, color=rsupports] {\typename{supports}} (knowledge);
      \draw[rel, color=rrefines] (knowledge)
        to[out=123, in=57, looseness=5.9]
        node[lbl, above, color=rrefines] {\typename{refines}} (knowledge);

    \end{tikzpicture}
    \caption{\textbf{GRACE network schema $T_G=(\mathcal{A},\mathcal{R},\{\sigma_r\}_{r\in\mathcal{R}})$.} Nodes
   denote object types in $\mathcal{A}$, and directed labeled edges denote admissible relation types:
  {\color{rsupports}\typename{supports}}, {\color{rrefines}\typename{refines}}, and
  {\color{rsequence}\typename{sequence}}.}
    \label{fig:network-schema}
  \end{figure}
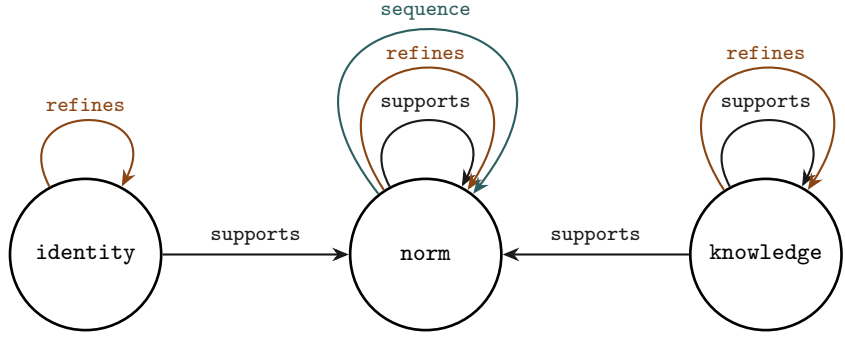

Following the formulation of heterogeneous information networks \citep{sun2011pathsim}, GRACE maintains the substrate as a directed typed graph $G_t=(V_t,E_t)$ at update step $t$, where nodes encode atomic instruction units and edges encode directed structural relations. The graph is equipped with an object-type mapping function $\varphi:V\rightarrow\mathcal{A}$ and a relation-type mapping function $\psi:E\rightarrow\mathcal{R}$. GRACE instantiates the object and relation type sets as
\begin{equation}\label{eq:type-sets}
\mathcal{A}=\{\mathsf{identity},\mathsf{norm},\mathsf{knowledge}\},
\qquad
\mathcal{R}=\{\mathsf{supports},\mathsf{refines},\mathsf{sequence}\}.
\end{equation}
Following the ontology-specification convention of \citet{gruber1993translation}, we specify each object and relation type by an identifier, a natural-language description of its intended meaning, and constraints that determine admissible instances. These definitions induce the network schema
\begin{equation}\label{eq:network-schema}
T_G=(\mathcal{A},\mathcal{R},\{\sigma_r\}_{r\in\mathcal{R}}),
\end{equation}
where each $\sigma_r\subseteq\mathcal{A}\times\mathcal{A}$ specifies the admissible source and target object types for relation type $r$. For every edge $e=(u,v)$ with $\psi(e)=r$, GRACE requires $(\varphi(u),\varphi(v))\in\sigma_r$, ensuring schema-conformant updates. For GRACE, the initial substrate $s_0$ is the graph $G_0$, obtained by a semantics-preserving decomposition of the shared initial instruction $\ell_0$ into schema-conformant objects and relations. Figure~\ref{fig:network-schema} visualizes the schema used in our implementation.

The object types correspond to classes in the Ontolingua convention, treated as unary relations whose instance variable ranges over admissible graph nodes. Relation-type definitions constrain source and target arguments and therefore determine the endpoint signatures in $\sigma_r$. Table~\ref{tab:type-definitions} summarizes the object and relation type definitions used by GRACE, and Table~\ref{tab:schema-matrix} gives the induced type-signature matrix.

\begin{table}[H]
\vspace{-0.5em}
\centering
\scriptsize
\setlength{\tabcolsep}{3pt}
\begin{tabular}{lll p{0.55\linewidth}}
\toprule
\textbf{Kind} & \textbf{Symbol} & \textbf{Role} & \textbf{Necessary condition} \\
\midrule
Object & $\mathsf{identity}$ & Agent standing & Content fixes the agent's mandate and the subject to which conduct is attributed. \\
Object & $\mathsf{norm}$ & Conduct rule & Content states a standard the agent's action is required to satisfy or breach. \\
Object & $\mathsf{knowledge}$ & Domain premise & Content states what is the case in the task domain and serves as a reasoning premise. \\
\midrule
Relation & $\mathsf{supports}$ & Grounding & $(\varphi(u),\varphi(v))\in\sigma_{\mathsf{supports}}$ and $u$ grounds, justifies, or reinforces $v$. \\
Relation & $\mathsf{refines}$ & Specialization & $\varphi(u)=\varphi(v)$; $u$ is a narrower case of $v$ obtained by adding conditions, exceptions, or operational detail without contradiction. \\
Relation & $\mathsf{sequence}$ & Procedure order & $\varphi(u)=\varphi(v)=\mathsf{norm}$ and $u$ precedes $v$ in procedural execution order. \\
\bottomrule
\end{tabular}
\caption{\textbf{Object and relation type definitions in the GRACE network schema.} Each row gives the natural-language rendering of the corresponding \texttt{:def} condition.}
\label{tab:type-definitions}
\vspace{-0.5em}
\end{table}

 % ===================================================================
  % fig/network_schema.tex — Network schema diagram for GRACE
  %
  % Required preamble in main.tex:
  %   \usepackage{tikz}
  %   \usetikzlibrary{arrows.meta}
  %   \definecolor{rsupports}{HTML}{1A1A1A}
  %   \definecolor{rrefines}{HTML}{8B4513}
  %   \definecolor{rsequence}{HTML}{2C5F5F}
  %   \newcommand{\typename}[1]{\texttt{#1}}
  %
  % Usage: \input{fig/network_schema}
  % ===================================================================

\begin{table}[H]
\vspace{-0.5em}
\centering
\scriptsize
\setlength{\tabcolsep}{4pt}
\begin{tabular}{lccc}
\toprule
\textbf{Source type} & \textbf{$\mathsf{identity}$} & \textbf{$\mathsf{norm}$} & \textbf{$\mathsf{knowledge}$} \\
\midrule
\textbf{$\mathsf{identity}$} & $\mathsf{refines}$ & $\mathsf{supports}$ & $\emptyset$ \\
\textbf{$\mathsf{norm}$} & $\emptyset$ & $\mathsf{supports}, \mathsf{refines}, \mathsf{sequence}$ & $\emptyset$ \\
\textbf{$\mathsf{knowledge}$} & $\emptyset$ & $\mathsf{supports}$ & $\mathsf{supports}, \mathsf{refines}$ \\
\bottomrule
\end{tabular}
\caption{\textbf{Type-signature matrix for the GRACE network schema.} Each cell lists relation types allowed from the source object type to the target object type.}
\label{tab:schema-matrix}
\vspace{-0.5em}
\end{table}

\paragraph{Coverage of the schema.}
\textbf{The question.} The three object types are deliberately coarse. What happens to instruction content that is not a role statement, a conduct rule, or a domain premise?

\textbf{What the decomposition is told to do.} Prompt-to-graph transcribes the instruction completely: omit nothing, invent nothing. It splits multi-step procedures, exceptions, conditional branches, and enumerations into independently checkable units, and it uses relations to retain grounding, specialization, and order (Appendix~\ref{app:prompt-p2g}).

\textbf{How the telecom instruction maps.} The role statement becomes the single $\mathsf{identity}$ node. Conduct rules, tool-usage rules (one tool call per turn, when to transfer to a human agent), verbatim message templates, and the individual steps of diagnostic and repair procedures become $\mathsf{norm}$ nodes; $\mathsf{sequence}$ edges carry procedural order and $\mathsf{refines}$ edges carry exceptions and special cases. Entity definitions (customer, line, plan, device, bill, payment method), the actions available on the user's device, and situational facts such as the current time become $\mathsf{knowledge}$ nodes that $\mathsf{support}$ the norms relying on them. Tool signatures are not part of the instruction in this harness; the benchmark passes them through the API, so the graph holds only the rules about their use.

\textbf{What has no dedicated type.} Few-shot examples, output-format templates, and remarks about the instruction itself are carried as $\mathsf{norm}$ or $\mathsf{knowledge}$ nodes with verbatim content. They are preserved and edited like any other unit, but verification is only as fine-grained as their typing allows.

\textbf{What the fidelity check guarantees.} After decomposition, a check reports missing units, distorted content, mistyped nodes, and wrong or missing relations against the source text (Appendix~\ref{app:prompt-p2g-fidelity}). The same LLM performs the decomposition and the check, so it guards against omission rather than certifying semantic completeness.

\textbf{The schema is configuration, not method.} Object and relation types, their definitions, admissible pairs, acyclicity constraints, and the grounding policy are loaded from a YAML specification or constructed in code. The public release ships an example schema with role, control, and procedure types and governs, implements, and precedes relations. The default ontology is a minimal vocabulary for conduct-governing instructions, not a claim that three types suffice for every instruction.

\subsection{Evolution Pipeline}\label{sec:pipeline}
At each update step, GRACE instantiates the evolution operator in Eq.~\ref{eq:evolve} as a mapping from the previous graph substrate, previous instruction checkpoint, and diagnostic observation to the next graph substrate and instruction checkpoint, $(G_{t-1},\ell_{t-1},o_t)\mapsto(G_t,\ell_t)$. The pipeline obtains these outputs through operation planning, structural validation, and delta reconstruction.

\paragraph{Operation planning.}
The first stage converts the diagnosis into schema-constrained graph updates. Let each graph state be represented as $(V,E,\varphi,\psi,m)$, where $m:V\rightarrow\mathcal{T}$ maps each node to its natural-language instruction content. GRACE defines a finite editing algebra over $T_G$:
\begin{equation}\label{eq:editing-algebra}
\mathcal{O}_{T_G}
=
\{
\mathrm{AddNode},
\mathrm{AddEdge},
\mathrm{ModifyNode},
\mathrm{RemoveNode},
\mathrm{RemoveEdge},
\mathrm{Merge}
\}.
\end{equation}
An operation belongs to $\mathcal{O}_{T_G}$ only when applying it preserves schema conformance under $T_G$. Table~\ref{tab:graph-editing-operators} summarizes the six operators. An LLM-mediated edit planner with a fixed prompt interface proposes a finite set of typed graph edits over this algebra:
\begin{equation}\label{eq:edit-plan}
\Delta_t^{\mathrm{edit}}
=
F_{\mathrm{edit}}(G_{t-1},o_t),
\qquad
\Delta_t^{\mathrm{edit}}\subseteq\mathcal{O}_{T_G}.
\end{equation}

\begin{table}[H]
\vspace{-0.5em}
\centering
\scriptsize
\setlength{\tabcolsep}{3.5pt}
\begin{tabular}{l l p{0.55\linewidth}}
\toprule
\textbf{Operator} & \textbf{Arguments} & \textbf{Description} \\
\midrule
$\mathrm{AddNode}$ & $(a,m_v)$ & Introduces node $v$ with $\varphi(v)=a$ and content $m(v)=m_v$, where $a\in\mathcal{A}$. \\
$\mathrm{AddEdge}$ & $(u,v,r)$ & Introduces edge $e=(u,v)$ with $\psi(e)=r$ when $r\in\mathcal{R}$ and $(\varphi(u),\varphi(v))\in\sigma_r$. \\
$\mathrm{ModifyNode}$ & $(v,m_v')$ & Replaces $m(v)$ with $m_v'$ while preserving $\varphi(v)$. \\
$\mathrm{RemoveNode}$ & $(v)$ & Removes $v$ and all incident edges from the graph. \\
$\mathrm{RemoveEdge}$ & $(e)$ & Removes an existing edge $e$ from the graph. \\
$\mathrm{Merge}$ & $(u,v,m_w)$ & For same-type nodes, replaces $u$ and $v$ with representative node $w$ and reroutes admissible incident edges under $T_G$. \\
\bottomrule
\end{tabular}
\caption{\textbf{Graph editing operators in $\mathcal{O}_{T_G}$.}}
\label{tab:graph-editing-operators}
\vspace{-0.5em}
\end{table}

Each edit in $\Delta_t^{\mathrm{edit}}$ is traceable to a diagnosis finding and must conform to $T_G$. For every proposed relation edge $e=(u,v)$ with $\psi(e)=r$, the source and target object types must satisfy $(\varphi(u),\varphi(v))\in\sigma_r$. The planned edits $\Delta_t^{\mathrm{edit}}$ are deterministically applied to $G_{t-1}$, yielding an intermediate graph $\widehat{G}_t$ together with the edit-touched node set $U_t^{\mathrm{edit}}$. The intermediate graph is then passed to a relation-maintenance module before structural validation. This LLM-mediated pass reviews the typed relation layer of $\widehat{G}_t$, focusing on the relations incident to the edit-touched set $U_t^{\mathrm{edit}}$, and proposes relation-level edits:
\begin{equation}\label{eq:relation-plan}
\Delta_t^{\mathrm{rel}}
=
F_{\mathrm{relation}}(\widehat{G}_t, U_t^{\mathrm{edit}}),
\qquad
\Delta_t^{\mathrm{rel}}\subseteq\mathcal{O}_{T_G}.
\end{equation}
Relation-level edits must also conform to $T_G$. The accepted relation edits $\Delta_t^{\mathrm{rel}}$ are deterministically incorporated into $\widehat{G}_t$, yielding the candidate graph $G_t^{\mathrm{cand}}$ together with the relation-touched set $U_t^{\mathrm{rel}}$.

\paragraph{Structural validation.}
The candidate graph $G_t^{\mathrm{cand}}$ is first checked for conformance to the network schema. Structural analysis (SA) then examines the affected typed neighborhoods for local semantic issues, and flagged issues are repaired through schema-conformant operations. The scoping rule borrows one principle from PathSim \citep{sun2011pathsim}: in a heterogeneous information network, the meaningful comparison context for an object is defined by typed paths around it, not by the whole network. GRACE applies the same idea to validation rather than to similarity ranking. A typed local neighborhood defines which instruction units are compared and which structural evidence can distinguish, merge, or repair them.

Let $U_t = U_t^{\mathrm{edit}} \cup U_t^{\mathrm{rel}}$ denote the update-touched node set, and let $d_G$ denote shortest-path distance in graph $G$. For any validation radius $k$, we define the $k$-hop typed neighborhood as
\begin{equation}\label{eq:validation-scope}
N_k(U_t,G_t^{\mathrm{cand}})
=
\left\{
 v\in V_t^{\mathrm{cand}}:
\min_{u\in U_t} d_{G_t^{\mathrm{cand}}}(u,v)\le k
\right\}.
\end{equation}
Here, $k$-hop refers to all nodes reachable from $U_t$ within at most $k$ graph hops. Distances are computed on the relation layer of the graph: edges incident to $\mathsf{identity}$ nodes are excluded from traversal, because every norm is grounded in the agent's identity through a $\mathsf{supports}$ edge and this grounding backbone would otherwise place the whole graph within two hops of any node. The exclusion is a schema-level setting (the grounding policy of the network schema), not a hard-coded rule, so a custom schema can name a different anchor type or none. In implementation, structural analysis uses a deterministic progressive schedule that starts from $k=3$ and increases the radius by 3 across a fixed number of local expansion rounds. Let $k^\star$ denote the final radius reached by this schedule. The validation neighborhood $H_t$ is the local typed subgraph induced by $N_{k^\star}(U_t,G_t^{\mathrm{cand}})$ inside $G_t^{\mathrm{cand}}$, containing the selected neighborhood nodes, all candidate-graph edges among them, and the inherited type mappings. Structural analysis is performed on $H_t$, and repairs are applied back to $G_t^{\mathrm{cand}}$ to obtain $G_t$.

Structural analysis detects two semantic issue families. First, contradiction captures joint unsatisfiability under overlapping applicability scope. For nodes $u,v\in V(H_t)$, let $\operatorname{Overlap}_{H_t}(u,v)$ indicate that the two units apply under overlapping graph-local scope, and let $\operatorname{Mod}_{H_t}(x)$ denote the set of coherent instruction interpretations satisfying content $x$ in the validation neighborhood $H_t$. We define
\begin{equation}\label{eq:contradiction}
\operatorname{Contr}_{H_t}(u,v)
\iff
\operatorname{Overlap}_{H_t}(u,v)
\land
\operatorname{Mod}_{H_t}\big(m(u)\land m(v)\big)=\emptyset .
\end{equation}
This follows the standard view that contradictory contents cannot be jointly satisfied, adapted to natural-language instruction units whose applicability is determined by the validation neighborhood \citep{demarneffe2008finding}.

Second, redundancy captures lack of marginal semantic contribution within the same object type. We first define a redundancy candidate:
\begin{equation}\label{eq:red-candidate}
\begin{aligned}
\operatorname{RedCand}_{H_t}(u,v)
\iff\;&
\varphi(u)=\varphi(v)
\land
\operatorname{Overlap}_{H_t}(u,v) \\
&\land
\left(
\operatorname{Entails}_{H_t,\varphi(u)}(u,v)
\lor
\operatorname{Entails}_{H_t,\varphi(u)}(v,u)
\right).
\end{aligned}
\end{equation}
The predicate $\operatorname{Entails}_{H_t,a}(u,v)$ denotes type-conditioned semantic subsumption for object type $a\in\mathcal{A}$: under the validation neighborhood $H_t$, the content of $v$ can be inferred from the content of $u$. This follows textual-entailment theory \citep{androutsopoulos2010survey} and operationalizes redundancy as informational equivalence or subsumption between texts \citep{zanzotto2011linguistic}. The same-type restriction prevents identity, norm, and knowledge nodes from being collapsed merely because their natural-language contents are related.
A candidate becomes an actionable redundancy issue only when the graph provides no structural reason to keep the two units distinct:
\begin{equation}\label{eq:redundancy}
\operatorname{Red}_{H_t}(u,v)
\iff
\operatorname{RedCand}_{H_t}(u,v)
\land
\neg\operatorname{Diff}_{H_t}(u,v).
\end{equation}
Here, $\operatorname{Diff}_{H_t}(u,v)$ indicates that the validation neighborhood structurally differentiates the two units. Differentiating evidence can include specialization, grounding, procedural ordering, or distinct typed support neighborhoods, preventing semantically related but functionally distinct instruction units from collapsing. Structural analysis flags pairs that satisfy either contradiction or actionable redundancy:
\begin{equation}\label{eq:issue-set}
\mathcal{I}_t
=
\left\{
(u,v)\in V(H_t)\times V(H_t):
\operatorname{Contr}_{H_t}(u,v)
\lor
\operatorname{Red}_{H_t}(u,v)
\right\}.
\end{equation}
Detected issues in $\mathcal{I}_t$ are repaired through schema-conformant operations in $\mathcal{O}_{T_G}$, resulting in the validated graph $G_t$.

\paragraph{Delta reconstruction}\label{sec:delta}
After validation, delta reconstruction converts the graph-level change log $\mathcal{L}_t^G = G_t \triangle G_{t-1}$ into the deployed instruction checkpoint, where $\triangle$ denotes the component-wise symmetric difference over the typed graph state, interpreted from $G_{t-1}$ to $G_t$. The instruction checkpoint is produced as $\ell_t=F_{\mathrm{recon}}(\ell_{t-1},\mathcal{L}_t^G)$, editing only text regions aligned with validated graph-level changes while preserving unaffected content, formatting, and section structure. At deployment, $\ell_t$ is assembled with harness-provided information and task input to form the agent context supplied to $\pi$.

\section{Experiments}\label{sec:exp}

\subsection{Experimental Setup}\label{sec:setup}

\paragraph{Benchmark, harness, and comparison scope.}
We evaluate on the telecom domain of $\tau^2$-bench \citep{barres2026tau2bench}, which comprises 2,285 composable tasks constructed from 15 atomic subtask groups across three intent categories. Within this harness, the evolving component is the persistent system-level instruction included in the agent context. All conditions share the same initial instruction, experience batches, diagnosis procedure, held-out evaluation set, agent model, and user simulator. They differ only in how $F_{\mathrm{Evolve}}$ maintains, validates, and reconstructs this persistent context component.

\paragraph{Controlled shift protocol and task allocation.}
To evaluate whether the evolved instruction component preserves reliable behavior when the experience distribution changes, we construct a controlled shift protocol over user intents. The protocol preserves subpopulation support across phases while changing relative intent frequencies, matching the subpopulation-shift setting of \citet{koh2021wilds}. We partition 486 tasks into a held-out evaluation set of 66 tasks and 10 experience batches of 42 tasks each, disjoint at the task level but sharing underlying atomic subtask groups. The experience batches alternate between two intent-mixture configurations across the 10-batch trajectory, as detailed in Table~\ref{tab:task-allocation}; Appendix~\ref{app:shift} quantifies the induced shift over intents and atomic subtask groups.

\begin{table}[t]
\vspace{-2em}
\begin{center}
\begin{tabular}{lcccc}
\toprule
& \textbf{MMS} & \textbf{Mobile Data} & \textbf{Service} & \textbf{Total} \\
\midrule
\textbf{Evaluation} & 30 (10/10/10) & 30 (10/10/10) & 6 (2/2/2) & 66 \\
\textbf{Exp. Phase~A} (B0,1,4,5,8,9) & 30 (10/10/10) & 9 (3/3/3) & 3 (1/1/1) & 42 \\
\textbf{Exp. Phase~B} (B2,3,6,7) & 9 (3/3/3) & 30 (10/10/10) & 3 (1/1/1) & 42 \\
\bottomrule
\end{tabular}
\end{center}
\caption{\textbf{Controlled shift protocol and task allocation.} The held-out evaluation set is shared across all checkpoints and never used for evolution. Experience batches alternate between two intent-mixture phases, both containing all three intents. MMS and Mobile Data are reweighted in opposite directions, while Service is held constant. Numbers in parentheses denote Easy/Hard/None customer persona splits. No two batches share any task.}\label{tab:task-allocation}
\end{table}

\paragraph{Evaluation metrics.}
At each reported evaluation checkpoint, the agent is evaluated using the deployed context assembled from instruction checkpoint $\ell_t$ on the same 66-task held-out set with three independent trials per task, yielding 198 episodes per checkpoint. The replicated evaluation reports the shared initial instruction $\ell_0$ and later checkpoints $\ell_6,\ell_8,\ell_{10}$. Following $\tau$-bench \citep{yao2024taubench}, pass@$k$ measures task-level capability under a best-of-$k$ criterion, while pass\textasciicircum{}$k$ measures trial-level reliability under an all-$k$ criterion. To measure retention under phase changes, we report intent-level backward transfer following \citet{lopezpaz2017gem}: for a phase transition from $X$ to $Y$ and an intent $d$ not emphasized in Phase~$Y$, $\text{BWT}_d(X \to Y) = \text{pass\textasciicircum{}}k_d(\ell_{Y_{\text{end}}}) - \text{pass\textasciicircum{}}k_d(\ell_{X_{\text{end}}})$. Positive BWT indicates retained or improved performance on that intent after the experience distribution shifts away from it. In all reported BWT results, $k=3$. The post-hoc consistency audit used for diagnostic analysis is described in Appendix~\ref{app:audit}.

\paragraph{Evolution conditions and shared diagnosis.}
We compare three primary instantiations of $F_{\mathrm{Evolve}}$. \textit{Holistic Context Evolution} (\textit{HCE}) applies a single-pass text-maintained update to the persistent instruction with a built-in consistency guard. \textit{GRACE without SA} ablates structural analysis from the graph-regularized instruction substrate, while \textit{GRACE} denotes the full method. These are controlled instantiations of the same evolution interface, designed to isolate the effects of representation and structural validation rather than to reimplement full prior systems with different artifacts and protocols. All conditions use Gemini 2.5 Flash as the agent model and GPT-4.1 as the user simulator. They share the same two-stage diagnosis procedure, which reflects on failed conversations and synthesizes the resulting findings into prioritized update themes. The procedure is fixed across conditions, but diagnosis content may differ after the first update. Implementation settings, the structural-analysis schedule, and every prompt are given in Appendices~\ref{app:impl} and~\ref{app:prompts}. The comparison matches the initial instruction, experience batches, diagnosis procedure, held-out evaluation set, and one batch-level evolution opportunity per update, but not internal LLM-call budgets. Zero-shot Gemini 2.5 Flash and Gemini 3.1 Pro references are included only as fixed held-out references; the evolution comparison itself is between the three Gemini 2.5 Flash conditions above.

\paragraph{Compute accounting.}
\textbf{Budgets are reported, not equalized.} The three conditions are not matched in internal LLM-call budget. Table~\ref{tab:evolution-cost} gives the offline evolution cost per batch from the per-call pricing ledger. GRACE makes 6.3 calls and reads about 241K tokens per evolution step. GRACE without SA makes 3.1 calls and reads 131K tokens. HCE makes a single call of about 39K tokens. GRACE also pays a one-time prompt-to-graph initialization of 16 calls (\$0.78 per run).

\textbf{Where the extra calls go.} Operation planning and relation maintenance read the whole graph. Up to three structural-analysis rounds read only the local subgraph.

\textbf{What the comparison can attribute.} The observed gap is attributable to the combination of representation, scoped verification, and the additional verification calls. It does not isolate the contribution of extra inference compute alone. A budget-matched flat-text variant is the natural control we have not run (Section~\ref{sec:limitations}).

\textbf{Evolution is a small share of the whole run.} Table~\ref{tab:cost-effectiveness} totals experience collection, diagnosis, evolution, and initialization per ten-batch run. GRACE spends \$42.59 against \$40.48 for HCE, a 5\% difference. Experience collection dominates every condition, and the failure-driven diagnosis stage is cheaper for the condition that fails less. Per dollar of offline spend, GRACE's pass\textasciicircum{}3 gain over the shared initial instruction is 5.4$\times$ that of HCE. The evolution-stage ratio of 4.4$\times$ is the number for attributing mechanism; the whole-run ratio of 1.05$\times$ is the number for judging affordability. The deployed instruction is also read on every agent turn, and Section~\ref{sec:analysis} shows that the deployment-time cost ordering is the reverse of the evolution-time ordering.

\begin{table}[t]
\centering
\small
\setlength{\tabcolsep}{4pt}
\begin{tabular}{lccccc}
\toprule
\textbf{Method} & \textbf{Calls/evol.} & \textbf{Tokens/evol.} & \textbf{Cost/evol.} & \textbf{Latency/evol.} & \textbf{Cost/run} \\
\midrule
GRACE & 6.3 & 241,058 & \$0.275 & 400s & \$2.75 \\
w/o SA & 3.1 & 130,560 & \$0.140 & 197s & \$1.40 \\
HCE & 1.0 & 39,427 & \$0.062 & 93s & \$0.62 \\
\bottomrule
\end{tabular}
\caption{\textbf{Offline evolution cost per batch} (mean over the four replications with per-call cost ledgers). Cost/run covers 10 evolution batches. GRACE additionally performs one prompt-to-graph initialization per run: 16 calls, 711,310 tokens, and \$0.78.}
\label{tab:evolution-cost}
\end{table}

\begin{table}[H]
\centering
\small
\begin{tabular}{lccccccc}
\toprule
\textbf{Condition} & \textbf{Experience} & \textbf{Diagnosis} & \textbf{Evolution} & \textbf{Init} & \textbf{Total} & $\boldsymbol{\Delta}$\textbf{pass\textasciicircum{}3} & \textbf{per \$} \\
\midrule
GRACE & \$36.02 & \$3.04 & \$2.75 & \$0.78 & \$42.59 & +0.570 & +0.0134 \\
GRACE w/o SA & \$33.24 & \$4.49 & \$1.40 & \$0.78 & \$39.91 & +0.157 & +0.0039 \\
HCE & \$34.25 & \$5.61 & \$0.62 & \$0.00 & \$40.48 & +0.100 & +0.0025 \\
\bottomrule
\end{tabular}
\caption{\textbf{Whole-run offline cost per condition and reliability gain per dollar.} Costs are per ten-batch run, mean over the four replications with per-call ledgers; held-out evaluation (\$52--\$55 per run, identical protocol cost for every condition) is excluded. Diagnosis is failure-driven, so cheaper for conditions that fail less; the ledger summary reports per-condition diagnosis cost for GRACE and HCE, and the pooled mean (\$4.49) is used for GRACE without SA. $\Delta$pass\textasciicircum{}3 is the mean $\ell_{10}$ value minus the shared $\ell_0$ value of 0.091.}
\label{tab:cost-effectiveness}
\end{table}

\paragraph{Replication and audit scope.}
Each primary condition is run for five independent evolution trajectories. At every reported checkpoint, each trajectory is evaluated on the same 66 held-out tasks with three trials per task; each method-checkpoint mean therefore aggregates five 198-episode evaluations. HCE receives the same diagnosis signal and update schedule as GRACE and maintains the deployed instruction directly as text with an explicit consistency guard, so it is not a zero-shot prompt baseline. The post-hoc LLM-as-Judge audit is applied only to $\ell_0,\ell_6,\ell_8,\ell_{10}$ and is used only for mechanism analysis. It is not used to select checkpoints, modify instruction checkpoints, or compute pass\textasciicircum{}3, pass@3, or pass@1. Appendix~\ref{app:shift} and Appendix~\ref{app:audit} give the controlled-shift quantification and the audit protocol.

\subsection{Main Results}\label{sec:results}

\begin{figure}[t]
\begin{center}
\includegraphics[width=\linewidth]{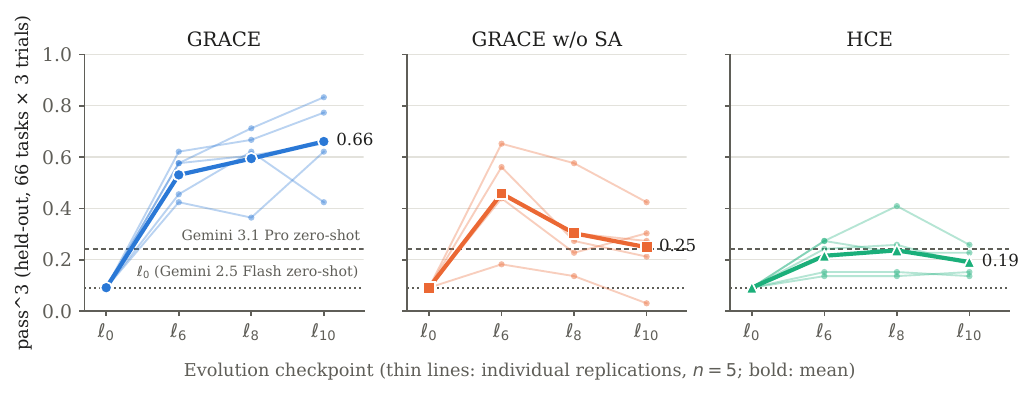}
\vspace{-0.6em}
\end{center}
\caption{\textbf{Per-replication pass\textasciicircum{}3 trajectories, one panel per condition.} Thin lines are the five independent evolution trajectories of each condition, evaluated at the shared initial checkpoint $\ell_0$ and at $\ell_6$, $\ell_8$, $\ell_{10}$ (198 held-out episodes per point); the bold line is their mean, with the $\ell_{10}$ mean printed at the right. Checkpoints are spaced categorically because only these four were evaluated. Dashed and dotted lines mark the Gemini 3.1 Pro and Gemini 2.5 Flash zero-shot references on the same held-out set. Every GRACE trajectory ends at or above every ablation trajectory and above every HCE trajectory; the ablation's individual runs peak at $\ell_6$ and then decline, and no HCE trajectory exceeds 0.41 at any evaluated checkpoint.}\label{fig:trajectory}
\end{figure}

Table~\ref{tab:main-results} summarizes the checkpoint results, and Figure~\ref{fig:trajectory} shows the replicated pass\textasciicircum{}3 trajectories. The conditions share the agent model, simulator, experience batches, diagnosis procedure, held-out set, and update schedule. The comparison therefore turns on how each instantiation of $F_{\mathrm{Evolve}}$ maintains, verifies, and reconstructs the evolving instruction component.

\begin{table}[H]
\centering
\scriptsize
\setlength{\tabcolsep}{3.5pt}
\begin{tabular}{llccc}
\toprule
\textbf{Metric} & \textbf{Method} & $\boldsymbol{\ell_6}$ & $\boldsymbol{\ell_8}$ & $\boldsymbol{\ell_{10}}$ \\
\midrule
\multirow{3}{*}{pass\textasciicircum{}3}
& GRACE & .530 (.086) & .594 (.135) & .661 (.158) \\
& w/o SA & .458 (.177) & .303 (.165) & .248 (.144) \\
& HCE & .215 (.066) & .236 (.109) & .191 (.051) \\
\midrule
\multirow{3}{*}{pass@3}
& GRACE & .942 (.075) & .936 (.071) & .979 (.025) \\
& w/o SA & .845 (.189) & .670 (.221) & .685 (.254) \\
& HCE & .618 (.218) & .418 (.286) & .479 (.244) \\
\midrule
\multirow{3}{*}{pass@1}
& GRACE & .757 (.078) & .775 (.109) & .851 (.073) \\
& w/o SA & .662 (.192) & .490 (.196) & .468 (.199) \\
& HCE & .400 (.116) & .327 (.201) & .330 (.138) \\
\bottomrule
\end{tabular}
\caption{\textbf{Replicated checkpoint summary.} Values are mean (std) over five runs. The shared initial instruction has pass\textasciicircum{}3 = .091, pass@3 = .333, and pass@1 = .202.}\label{tab:main-results}
\end{table}

\paragraph{GRACE sustains cumulative improvement.}
Across five independent replications, GRACE is the only primary condition that shows sustained improvement through the later checkpoints of the controlled shift protocol (Figure~\ref{fig:trajectory}). Starting from the shared initial instruction with pass\textasciicircum{}3 of 0.091, it reaches 0.530$\pm$0.086 at $\ell_6$, 0.594$\pm$0.135 at $\ell_8$, and 0.661$\pm$0.158 at $\ell_{10}$. Its pass@3 remains near ceiling over the same checkpoints, reaching 0.979$\pm$0.025 at $\ell_{10}$ (Table~\ref{tab:main-results}). The remaining gap between pass@3 and pass\textasciicircum{}3 indicates that capability is easier to recover than strict reliability, which requires all three independent trials to succeed. Table~\ref{tab:frontier} puts the gain in perspective: with no evolution, the strongest frontier model we evaluated on the same held-out set, Gemini 3.1 Pro, reaches pass\textasciicircum{}3 of 0.242, and GPT-5.1 reaches 0.121; the evolved Gemini 2.5 Flash agent ends above both in every one of its five trajectories.

\begin{table}[t]
\centering
\small
\begin{tabular}{lccc}
\toprule
\textbf{Model} & \textbf{pass\textasciicircum{}3} & \textbf{pass@3} & \textbf{pass@1} \\
\midrule
Gemini 2.5 Flash (agent model) & 0.091 & 0.333 & 0.202 \\
Gemini 2.5 Pro & 0.121 & 0.455 & 0.293 \\
Gemini 3 Flash & 0.212 & 0.455 & 0.318 \\
Gemini 3.1 Pro & 0.242 & 0.742 & 0.485 \\
GPT-5.1 & 0.121 & 0.530 & 0.308 \\
\bottomrule
\end{tabular}
\caption{\textbf{Zero-shot frontier references on the same 66-task held-out set} with the default telecom instruction and no evolution. Gemini 2.5 Flash is the shared initial checkpoint $\ell_0$. The evolved Gemini 2.5 Flash agent under GRACE reaches pass\textasciicircum{}3 of 0.661$\pm$0.158 at $\ell_{10}$ (Table~\ref{tab:main-results}), above every zero-shot frontier model listed.}
\label{tab:frontier}
\end{table}

\paragraph{Flat-text instruction maintenance exhibits distinct failure patterns.}
HCE improves over the shared initial instruction but does not sustain the gain. It reaches 0.215$\pm$0.066 at $\ell_6$ and 0.236$\pm$0.109 at $\ell_8$, then finishes at 0.191$\pm$0.051. The same pattern is visible in pass@3: HCE reaches 0.618$\pm$0.218 at $\ell_6$ but drops to 0.479$\pm$0.244 at $\ell_{10}$. This trajectory illustrates that flat-text context evolution can exploit diagnostic signal but remains vulnerable to later degradation as the maintained instruction grows.

\paragraph{Graph structure alone is not sufficient for sustained improvement.}
GRACE without SA improves to 0.458$\pm$0.177 at $\ell_6$, but the gain is not consolidated. Its pass\textasciicircum{}3 drops to 0.303$\pm$0.165 at $\ell_8$ and 0.248$\pm$0.144 at $\ell_{10}$, while pass@3 falls from 0.845$\pm$0.189 to 0.685$\pm$0.254. The graph substrate makes the instruction artifact more verifiable, but the ablation shows that structure alone does not protect accumulated gains over time.

\paragraph{Replication-level statistics.}
\textbf{The trajectory is the unit of analysis.} Each condition has five independent trajectories, so we test on the per-replication pass\textasciicircum{}3 means (Table~\ref{tab:stats}).

\textbf{GRACE versus HCE.} The gap is significant at every reported checkpoint under Welch's $t$-test, an exact permutation test on the difference of means, and a bootstrap confidence interval that excludes zero: $+0.315$ at $\ell_6$ (95\% CI $[0.230, 0.400]$), $+0.358$ at $\ell_8$, and $+0.470$ at $\ell_{10}$ (95\% CI $[0.336, 0.594]$). Cohen's $d$ lies between 2.9 and 4.1.

\textbf{GRACE versus its ablation.} The gap is not distinguishable at $\ell_6$ ($+0.073$, permutation $p=0.54$). It separates at $\ell_8$ ($+0.291$, $p=0.016$) and $\ell_{10}$ ($+0.412$, $p=0.016$). This is the pattern expected if structural analysis matters through consolidation over time rather than through an immediate boost.

\textbf{Within-method trends.} Paired comparisons of $\ell_6$ and $\ell_{10}$ over the same five trajectories point the same way, although the GRACE increase is only marginally significant. GRACE improves by $+0.130$ (paired $t$, $p=0.06$). GRACE without SA declines by $-0.209$ ($p=0.005$). HCE is flat ($-0.024$, $p=0.14$). At $\ell_{10}$, all five GRACE trajectories exceed the Gemini 3.1 Pro zero-shot reference of 0.242 (minimum 0.424), against three of five for the ablation and one of five for HCE.

\textbf{What the tests do and do not show.} With $n=5$ the tests have limited power and the intervals are wide. We report them to show that the ordering is not an artifact of a single favorable run, not to claim precise effect magnitudes. All per-replication values are listed in Appendix~\ref{app:raw}. The same pattern holds for pass@3 and pass@1 (Table~\ref{tab:stats-other}): GRACE separates from HCE at every checkpoint on both metrics, and from its ablation only from $\ell_8$ onward.

\begin{table}[H]
\centering
\scriptsize
\setlength{\tabcolsep}{3pt}
\begin{tabular}{llccccc}
\toprule
\textbf{Checkpoint} & \textbf{Comparison} & \textbf{$\Delta$ mean} & \textbf{95\% bootstrap CI} & \textbf{Welch $p$} & \textbf{Permutation $p$} & \textbf{Cohen's $d$} \\
\midrule
$\ell_{6}$ & GRACE vs.\ HCE & +0.315 & [+0.230, +0.400] & $<$0.001 & 0.008 & 4.11 \\
$\ell_{6}$ & GRACE vs.\ GRACE w/o SA & +0.073 & [-0.073, +0.233] & 0.441 & 0.540 & 0.52 \\
\midrule
$\ell_{8}$ & GRACE vs.\ HCE & +0.358 & [+0.212, +0.482] & 0.002 & 0.016 & 2.91 \\
$\ell_{8}$ & GRACE vs.\ GRACE w/o SA & +0.291 & [+0.106, +0.443] & 0.017 & 0.016 & 1.93 \\
\midrule
$\ell_{10}$ & GRACE vs.\ HCE & +0.470 & [+0.336, +0.594] & 0.002 & 0.008 & 4.00 \\
$\ell_{10}$ & GRACE vs.\ GRACE w/o SA & +0.412 & [+0.245, +0.579] & 0.003 & 0.016 & 2.72 \\
\bottomrule
\end{tabular}
\caption{\textbf{Replication-level comparison of pass\textasciicircum{}3.} Each condition has five independent evolution trajectories ($n=5$ per group). Welch's $t$-test and an exact permutation test on the difference of means (252 splits) are two-sided; the bootstrap CI uses 20{,}000 resamples of the per-replication means.}
\label{tab:stats}
\end{table}

\begin{table}[H]
\centering
\scriptsize
\setlength{\tabcolsep}{2pt}
\begin{tabular}{p{1.1cm}p{2.6cm}ccc}
\toprule
\textbf{Metric} & \textbf{Comparison} & $\boldsymbol{\ell_6}$ & $\boldsymbol{\ell_8}$ & $\boldsymbol{\ell_{10}}$ \\
\midrule
pass@3 & GRACE vs.\ HCE & +0.324 [+0.15, +0.51], $p$=0.024 & +0.518 [+0.26, +0.71], $p$=0.024 & +0.500 [+0.28, +0.67], $p$=0.008 \\
pass@3 & GRACE vs.\ GRACE w/o SA & +0.097 [-0.05, +0.27], $p$=0.373 & +0.267 [+0.09, +0.45], $p$=0.040 & +0.294 [+0.10, +0.49], $p$=0.016 \\
\midrule
pass@1 & GRACE vs.\ HCE & +0.357 [+0.25, +0.47], $p$=0.008 & +0.447 [+0.25, +0.61], $p$=0.016 & +0.520 [+0.39, +0.63], $p$=0.008 \\
pass@1 & GRACE vs.\ GRACE w/o SA & +0.095 [-0.05, +0.27], $p$=0.365 & +0.284 [+0.10, +0.45], $p$=0.032 & +0.383 [+0.23, +0.55], $p$=0.008 \\
\bottomrule
\end{tabular}
\caption{\textbf{Replication-level comparison of pass@3 and pass@1.} Cells give the mean difference over five trajectories, a 95\% bootstrap interval, and the two-sided exact permutation $p$-value; pass\textasciicircum{}3 is in Table~\ref{tab:stats}.}
\label{tab:stats-other}
\end{table}

\subsection{Analysis}\label{sec:analysis}

The three trajectories differ not only in final score but also in failure mode. We next examine the mechanisms behind those differences.

\paragraph{Contradiction accumulation distinguishes HCE from the other failure modes.}
We use the post-hoc LLM-as-Judge audit described in \S\ref{sec:setup} to examine whether evolved instruction artifacts accumulate internal contradictions. The audit uses a single contradiction category. Two instruction statements are counted as contradictory when they prescribe incompatible behavior under overlapping conditions and the graph or text provides no structural relation that resolves the tension. Across the audited checkpoints $\ell_0,\ell_6,\ell_8,\ell_{10}$, HCE has the highest contradiction count at 3.35 on average, while GRACE and GRACE without SA remain in a narrower range from 1.75 to 1.90. Like the node counts and redundancy counts below, these audit statistics were computed on one complete trajectory per condition in the initial study and are not averaged over the five replications. This evidence is correlational, but it matches the observed performance collapse: as HCE's flat text grows, its built-in consistency guard has increasing difficulty keeping incompatible rules apart.

GRACE without SA keeps contradiction counts low but still fails to sustain improvement. Avoiding contradictions is therefore not enough; the remaining failure mode is lack of consolidation over the growing graph substrate.

\paragraph{Graph structure suppresses contradictions but does not consolidate.}
GRACE and GRACE without SA maintain similarly low contradiction counts, indicating that typed graph representation and incremental delta reconstruction already help avoid gross contradictions. The difference appears in growth and consolidation. By the final checkpoint, the graph-only ablation reaches 421 nodes, compared to 370 nodes for GRACE. The deployed instruction-size measurements in Table~\ref{tab:policy-size} show the same pattern at the text layer on the reported checkpoints: HCE grows substantially more than either graph-based condition, while GRACE remains slightly smaller than the graph-only ablation at $\ell_{10}$. The LLM-as-Judge redundancy audit points in the same direction. Across $\ell_0,\ell_6,\ell_8,\ell_{10}$, GRACE without SA has a higher redundancy count than GRACE on average, 4.85 compared to 4.15, and reaches 6.5 at $\ell_{10}$. This expansion matches the stochastic trajectory in \S\ref{sec:results}: individual checkpoints can improve, but the gains are not consolidated into a durable instruction component. The graph and SA play different roles. The graph makes relationships among accumulated instruction units explicit and helps prevent gross contradictions. SA consolidates the growing substrate by merging overlapping content and resolving subtle tensions.

\begin{table}[t]
\vspace{-0.5em}
\centering
\scriptsize
\setlength{\tabcolsep}{4pt}
\begin{tabular}{lcccc}
\toprule
\textbf{Method} & $\boldsymbol{\ell_0}$ & $\boldsymbol{\ell_6}$ & $\boldsymbol{\ell_8}$ & $\boldsymbol{\ell_{10}}$ \\
\midrule
GRACE & 23,363 & 42,902 & 46,471 & 51,413 \\
w/o SA & 23,363 & 45,207 & 49,343 & 54,893 \\
HCE & 23,363 & 59,668 & 73,822 & 82,592 \\
\bottomrule
\end{tabular}
\caption{\textbf{Mean deployed instruction size in characters.} Values average over five runs and measure the persistent system-level instruction assembled into the deployed agent context, not the internal graph substrate.}
\label{tab:policy-size}
\vspace{-0.5em}
\end{table}

\paragraph{Deployment-time cost reverses the evolution-time ordering.}
\textbf{Instruction size is a per-turn cost.} The deployed instruction is prepended to every agent turn, so Table~\ref{tab:policy-size} has a direct operational reading. At four characters per token, the mean $\ell_{10}$ instruction is about 12.9K tokens for GRACE, 13.7K for GRACE without SA, and 20.6K for HCE. The flat-text instruction therefore costs $1.6\times$ more input tokens on every inference call by the end of the trajectory. The ratio grows from $1.1\times$ at $\ell_1$ to $1.6\times$ at $\ell_{10}$.

\textbf{The separation holds per replication.} The smallest HCE $\ell_{10}$ (65.5K characters) is larger than the largest GRACE $\ell_{10}$ (61.5K). Growth is close to linear for every condition (least-squares $R^2$ of 0.978, 0.985, and 0.996 over $\ell_1$ to $\ell_{10}$): about 2.4K characters per update for GRACE, 2.9K for GRACE without SA, and 5.9K for HCE. The gap widens with every batch rather than saturating.

\textbf{Break-even.} At Gemini 2.5 Flash list input pricing at the time of writing (\$0.30 per million tokens), the 7.8K-token difference is about \$2.3 per thousand agent turns. GRACE's additional offline evolution and initialization cost of \$2.91 per run (Table~\ref{tab:evolution-cost}) is recovered after roughly 1{,}200 agent turns; for scale, one 42-episode experience batch produced about 1{,}760 agent and user-simulator calls in our ledger, so break-even falls within the first one or two batches of deployment. Consolidation pays for its own verification overhead at deployment scale. The open compute question is attribution, not affordability.

\paragraph{Retention under distribution shift reflects these mechanism differences.}
The controlled shift protocol tests whether these mechanism differences translate into retention under changing experience distributions. A \emph{cycle} is one shift-and-return: Cycle~1 runs from Phase~A (B0--B1) through Phase~B (B2--B3) back to Phase~A (B4--B5), and Cycle~2 from B4--B5 through B6--B7 to B8--B9; backward transfer for an intent is the change in its pass\textasciicircum{}3 between the ends of consecutive same-phase rounds. These values come from the initial single-trajectory study, in which every checkpoint of one trajectory per condition was evaluated; the five replications were evaluated only at $\ell_0,\ell_6,\ell_8,\ell_{10}$ and cannot reproduce them, so we report the pattern as illustrative rather than replicated. GRACE shows positive or zero backward transfer in all cycles except Mobile Data in Cycle~1, where a temporary drop of 0.233 is later recovered. HCE exhibits persistent negative backward transfer, with Mobile Data BWT reaching $-0.500$ in Cycle~1 and never recovering. The mechanism analysis explains the split. HCE's contradiction-laden instruction artifact loses coherence when the experience distribution shifts, while GRACE's consolidated instruction component is more resilient to the shift.

\section{Scope of the Evidence}\label{sec:limitations}

Three boundaries govern how the results should be read. Appendix~\ref{app:limitations} elaborates each.

\paragraph{One setting.}
All evidence comes from one domain, one agent model, and one harness: the telecom domain of $\tau^2$-bench with Gemini 2.5 Flash. We chose it because it is the benchmark's dual-control domain, where instruction quality has direct behavioral consequences. The results do not license claims about other domains, models, or harnesses. Coverage of the three-type schema has been examined only on this instruction.

\paragraph{Unmatched internal compute.}
The conditions share diagnosis and update schedule but not internal LLM-call budget. The experiments show that the graph-regularized operator, verification calls included, sustains improvement where a one-call flat-text operator does not. They do not show how much of the gap would survive a budget-matched flat-text control. The ablation spends half of GRACE's budget and still collapses, which argues against a pure compute explanation, but it does not settle the question.

\paragraph{Correlational mechanism evidence.}
Contradiction and redundancy counts come from a post-hoc LLM-judged audit. They support the mechanism story but never enter the headline metrics. With five trajectories per condition, the tests establish the ordering of the conditions at the later checkpoints, not precise effect sizes.

\section{Conclusion}\label{sec:conclusion}

This paper studied long-horizon agentic context evolution through the lens of verification. In the setting considered here, the evolving object is the persistent system-level instruction assembled into the agent context, while the model, tools, and harness remain fixed. GRACE maintains this instruction component as a typed semantic graph, enabling structural validation to focus on affected typed neighborhoods without rereading the entire flat-text artifact at each update. Under a controlled shift protocol across 10 evolution batches, replicated evaluation shows that GRACE sustains later-checkpoint improvement more reliably than both HCE and GRACE without SA. The ablation clarifies the mechanism. Contradiction avoidance helps, but durable improvement also requires active consolidation of the growing instruction substrate.

The evidence is limited in the ways Section~\ref{sec:limitations} describes, and the most informative next experiments are a budget-matched flat-text control, a second benchmark domain, and an audit with an independent judge. Within the controlled substrate-level setting studied here, the main lesson is direct. Long-horizon context evolution is easier to verify when relationships among accumulated instruction units are explicit and local enough to inspect.

\bibliographystyle{tmlr}
\bibliography{ref}

\newpage
\appendix

\section{Controlled Shift Protocol Quantification}\label{app:shift}

This appendix quantifies the controlled shift protocol used in the main experimental setup. The protocol changes the experience distribution over user intents while preserving the support of the intent subpopulations. This matches the subpopulation-shift setting of \citet{koh2021wilds}, where the same subpopulations are observed across distributions but their relative frequencies differ.

\paragraph{Protocol.}
The experience trajectory consists of 10 batches. Phase~A contains batches B0, B1, B4, B5, B8, and B9. Phase~B contains batches B2, B3, B6, and B7. Each batch contains 42 tasks. Both phases contain all three user intents. MMS and Mobile Data are reweighted in opposite directions, while Service is held constant.

\paragraph{Sample space and empirical distributions.}
Let $\mathcal{I}=\{M,D,S\}$ denote the user-intent set, corresponding to \texttt{mms\_issue}, \texttt{mobile\_data\_issue}, and \texttt{service\_issue}. Let $\mathcal{G}$ denote the 15 atomic subtask groups in the $\tau^2$-bench telecom domain \citep{barres2026tau2bench}. For each phase $\alpha\in\{A,B\}$, let $b_\alpha(i)$ be the per-batch count of intent $i$, $n_\alpha(i)$ the phase-aggregated count of intent $i$, and $m_\alpha(g,i)$ the number of occurrences of subtask group $g\in\mathcal{G}$ among tasks of intent $i$. Since one task may instantiate multiple subtask groups, subtask-group counts are occurrence counts. We use normalized counts as empirical distributions: $P_\alpha(i)=b_\alpha(i)/42$, $P_\alpha(g)=\sum_i m_\alpha(g,i)/\sum_{g',i}m_\alpha(g',i)$, and $P_\alpha(g\mid i)=m_\alpha(g,i)/\sum_{g'}m_\alpha(g',i)$.

\paragraph{Distance measures.}
For finite distributions $P$ and $Q$ over the same sample space $\mathcal{Z}$, we report total variation distance $\mathrm{TV}(P,Q)=\frac{1}{2}\sum_{z\in\mathcal{Z}}|P(z)-Q(z)|$ and KL divergence $D_{\mathrm{KL}}(P\|Q)=\sum_{z\in\mathcal{Z}}P(z)\log(P(z)/Q(z))$. KL values use natural logarithms. TV is symmetric. KL is asymmetric and finite only when the support of $P$ is contained in the support of $Q$.

\paragraph{Intent-level shift.}
Table~\ref{tab:app-intent-shift} reports both per-batch and phase-aggregated intent counts. The empirical intent distributions $P_A$ and $P_B$ are computed from the per-batch counts. Aggregating across all batches in the same phase gives the same distributions because every batch in a phase has the same intent composition.

\begin{table}[H]
\centering
\small
\begin{tabular}{lcccc}
\toprule
\textbf{Phase} & \textbf{MMS} & \textbf{Mobile Data} & \textbf{Service} & \textbf{Total} \\
\midrule
Phase~A per batch & 30 & 9 & 3 & 42 \\
Phase~A total & 180 & 54 & 18 & 252 \\
Phase~B per batch & 9 & 30 & 3 & 42 \\
Phase~B total & 36 & 120 & 12 & 168 \\
\bottomrule
\end{tabular}
\caption{Intent composition of the controlled shift protocol. Phase-level totals aggregate over six Phase~A batches and four Phase~B batches. The per-batch intent proportions are identical to the phase-aggregated intent proportions within each phase.}
\label{tab:app-intent-shift}
\end{table}

The empirical intent distributions are therefore
\[
P_A=\left(\frac{30}{42},\frac{9}{42},\frac{3}{42}\right),
\qquad
P_B=\left(\frac{9}{42},\frac{30}{42},\frac{3}{42}\right),
\]
over $(M,D,S)$. The resulting total variation distance is $\mathrm{TV}(P_A,P_B)=\tfrac{1}{2}\big(\tfrac{21}{42}+\tfrac{21}{42}+0\big)=0.500$, and the KL divergence is $D_{\mathrm{KL}}(P_A\|P_B)=\tfrac{30}{42}\log\tfrac{10}{3}+\tfrac{9}{42}\log\tfrac{3}{10}\approx 0.602$ (natural logarithm).

\paragraph{Subtask-level shift.}
Each task id lists the atomic subtasks it composes, so the same distances can be computed over atomic subtask identifiers using occurrence counts (a task with four atoms contributes four occurrences). The frozen split contains 19 distinct identifiers, several of which are variants of the same generator group; Phase~A batches contain 1{,}180 occurrences and Phase~B batches 659. Over this sample space, $\mathrm{TV}(P_A(g),P_B(g))=0.246$ and $D_{\mathrm{KL}}(P_A(g)\|P_B(g))\approx 0.261$, and every identifier has positive mass in both phases, so the support is preserved at the subtask level as well. The shift is carried almost entirely by the intent mixture rather than by within-intent composition: conditioned on intent, the total variation between phases is 0.062 for MMS, 0.091 for Mobile Data, and 0.188 for Service (the last over only 46 and 37 occurrences). The largest per-identifier changes are \texttt{bad\_vpn} (0.021 to 0.097) and \texttt{data\_saver\_mode\_on} (0.025 to 0.083), which rise in Phase~B, and \texttt{unseat\_sim\_card}, \texttt{break\_apn\_mms\_setting}, and \texttt{bad\_wifi\_calling}, which fall from roughly 0.08 to 0.02--0.04. These statistics are computed from \texttt{data/splits.json} in the released code by \texttt{paper\_data/subtask\_shift.py}.

\section{Audit Protocol}
\label{app:audit}
We use a fixed post-hoc audit rubric to inspect the audited checkpoints $\ell_0,\ell_6,\ell_8,\ell_{10}$ for contradiction, redundancy, and retention signals. The audit is applied after evolution and is not used to modify the instruction during the main trajectory. The audit therefore serves only as diagnostic evidence for interpreting the main trajectories; it is not part of $F_{\mathrm{Evolve}}$ for any condition.

For contradiction analysis, the judge receives the evolved instruction text and identifies pairs of instruction statements that prescribe incompatible behavior under overlapping conditions, unless the graph or text contains an explicit relation that resolves the apparent tension. For redundancy analysis, the judge identifies same-role instruction units that are semantically overlapping and lack a structural reason to remain separate. For retention analysis, we compute intent-level backward transfer using pass\textasciicircum{}3. Frontier LLM judges evaluate each audited checkpoint independently in two trials each, and the reported counts are medians across judges. Appendix~\ref{app:limitations} discusses the circularity risk when a judge shares the evolution operator's model family. Because these diagnostics are post-hoc and partially judgment-based, we treat them as correlational support for the mechanism analysis rather than as causal proof.

\section{Mechanism Audit Details}
\label{app:graph-size}
The initial prompt-to-graph conversion produces the same initial graph across repeated temperature-zero constructions, with 335 nodes and 442 edges. Node counts at the final checkpoint and the contradiction and redundancy counts over the audited checkpoints are reported in Section~\ref{sec:analysis}; the audit rubric and aggregation are given in Appendix~\ref{app:audit}.

\section{Cost Accounting}
\label{app:cost-accounting}
The costs in Tables~\ref{tab:evolution-cost} and~\ref{tab:cost-effectiveness} average over the four replications with per-call cost ledgers; the fifth replication was run on earlier infrastructure without a comparable ledger. Tokens combine input and output tokens, dollar costs come from the per-call pricing ledger for the agent and user-simulator models, and the latency column sums individual LLM-call latencies, so it is not wall-clock runtime because calls are partly parallelized.

\section{Per-Replication Results}\label{app:raw}
Tables~\ref{tab:raw-results} and~\ref{tab:raw-size} list every value behind the means and tests in the main text, so the statistics in Tables~\ref{tab:stats} and~\ref{tab:stats-other} can be recomputed from this appendix alone.

\begin{table}[H]
\centering
\scriptsize
\setlength{\tabcolsep}{4pt}
\resizebox{\linewidth}{!}{%
\begin{tabular}{llcccccccccc}
\toprule
& & & \multicolumn{3}{c}{\textbf{pass\textasciicircum{}3}} & \multicolumn{3}{c}{\textbf{pass@3}} & \multicolumn{3}{c}{\textbf{pass@1}} \\
\cmidrule(lr){4-6}\cmidrule(lr){7-9}\cmidrule(lr){10-12}
\textbf{Condition} & \textbf{Rep.} & & $\ell_6$ & $\ell_8$ & $\ell_{10}$ & $\ell_6$ & $\ell_8$ & $\ell_{10}$ & $\ell_6$ & $\ell_8$ & $\ell_{10}$ \\
\midrule
GRACE & \#1 & & 0.621 & 0.667 & 0.773 & 1.000 & 0.970 & 0.985 & 0.848 & 0.838 & 0.889 \\
 & \#2 & & 0.455 & 0.621 & 0.424 & 0.985 & 1.000 & 0.939 & 0.742 & 0.828 & 0.737 \\
 & \#3 & & 0.576 & 0.606 & 0.652 & 0.985 & 0.894 & 1.000 & 0.793 & 0.747 & 0.854 \\
 & \#4 & & 0.424 & 0.364 & 0.621 & 0.818 & 0.833 & 0.970 & 0.636 & 0.596 & 0.838 \\
 & \#5 & & 0.576 & 0.712 & 0.833 & 0.924 & 0.985 & 1.000 & 0.763 & 0.864 & 0.934 \\
& \textit{mean (sd)} & & \textbf{0.530} (0.086) & \textbf{0.594} (0.135) & \textbf{0.661} (0.158) & \textbf{0.942} (0.075) & \textbf{0.936} (0.071) & \textbf{0.979} (0.025) & \textbf{0.756} (0.078) & \textbf{0.775} (0.109) & \textbf{0.850} (0.073) \\
\midrule
GRACE w/o SA & \#1 & & 0.455 & 0.303 & 0.273 & 0.773 & 0.682 & 0.727 & 0.616 & 0.465 & 0.525 \\
 & \#2 & & 0.439 & 0.227 & 0.303 & 0.985 & 0.848 & 0.894 & 0.737 & 0.561 & 0.581 \\
 & \#3 & & 0.182 & 0.136 & 0.030 & 0.545 & 0.439 & 0.379 & 0.348 & 0.268 & 0.187 \\
 & \#4 & & 0.652 & 0.576 & 0.424 & 0.985 & 0.924 & 0.955 & 0.833 & 0.783 & 0.692 \\
 & \#5 & & 0.561 & 0.273 & 0.212 & 0.939 & 0.455 & 0.470 & 0.773 & 0.374 & 0.354 \\
& \textit{mean (sd)} & & \textbf{0.458} (0.177) & \textbf{0.303} (0.165) & \textbf{0.248} (0.144) & \textbf{0.845} (0.189) & \textbf{0.670} (0.221) & \textbf{0.685} (0.254) & \textbf{0.661} (0.192) & \textbf{0.490} (0.196) & \textbf{0.468} (0.199) \\
\midrule
HCE & \#1 & & 0.273 & 0.409 & 0.258 & 0.818 & 0.909 & 0.879 & 0.530 & 0.667 & 0.551 \\
 & \#2 & & 0.273 & 0.227 & 0.227 & 0.652 & 0.394 & 0.455 & 0.455 & 0.308 & 0.354 \\
 & \#3 & & 0.242 & 0.258 & 0.182 & 0.424 & 0.348 & 0.439 & 0.328 & 0.308 & 0.298 \\
 & \#4 & & 0.136 & 0.136 & 0.152 & 0.833 & 0.182 & 0.212 & 0.449 & 0.162 & 0.182 \\
 & \#5 & & 0.152 & 0.152 & 0.136 & 0.364 & 0.258 & 0.409 & 0.237 & 0.192 & 0.268 \\
& \textit{mean (sd)} & & \textbf{0.215} (0.066) & \textbf{0.236} (0.109) & \textbf{0.191} (0.051) & \textbf{0.618} (0.218) & \textbf{0.418} (0.286) & \textbf{0.479} (0.244) & \textbf{0.400} (0.116) & \textbf{0.327} (0.201) & \textbf{0.331} (0.138) \\
\bottomrule
\end{tabular}}

\caption{\textbf{Per-replication held-out results.} Every cell is one 198-episode evaluation (66 tasks $\times$ 3 trials) of one independent evolution trajectory. The shared initial instruction $\ell_0$ scores pass\textasciicircum{}3 = 0.091, pass@3 = 0.333, pass@1 = 0.202 (single shared measurement). GRACE \#2 at $\ell_{10}$ was evaluated three times (pass\textasciicircum{}3 = 0.409 / 0.485 / 0.379; every other cell once); the table reports their mean, 0.424, and their spread (sd 0.055) is the evaluation-to-evaluation noise floor for a 198-episode checkpoint. Its pass@3 and pass@1 come from the 0.485 evaluation, the only one for which these metrics were recorded. The means are the inputs to Tables~\ref{tab:stats} and~\ref{tab:stats-other}; Table~\ref{tab:main-results} is computed from unrounded episode counts, so a few of its entries differ from the means of these three-decimal values in the last digit.}
\label{tab:raw-results}
\end{table}

\begin{table}[H]
\centering
\scriptsize
\setlength{\tabcolsep}{3pt}
\resizebox{\linewidth}{!}{%
\begin{tabular}{llrrrrrrrrrrr}
\toprule
\textbf{Condition} & \textbf{Rep.} & $\ell_{0}$ & $\ell_{1}$ & $\ell_{2}$ & $\ell_{3}$ & $\ell_{4}$ & $\ell_{5}$ & $\ell_{6}$ & $\ell_{7}$ & $\ell_{8}$ & $\ell_{9}$ & $\ell_{10}$ \\
\midrule
GRACE & \#1 & 23,363 & 30,781 & 35,489 & 36,584 & 36,716 & 38,347 & 38,820 & 42,100 & 42,914 & 44,275 & 45,631 \\
 & \#2 & 23,363 & 27,363 & 34,454 & 42,644 & 45,884 & 50,232 & 52,321 & 54,503 & 56,015 & 58,578 & 61,478 \\
 & \#3 & 23,363 & 29,109 & 31,043 & 36,331 & 37,735 & 38,786 & 41,115 & 42,213 & 42,232 & 46,311 & 48,230 \\
 & \#4 & 23,363 & 25,906 & 30,026 & 34,729 & 36,233 & 39,836 & 43,155 & 47,381 & 48,418 & 50,822 & 52,063 \\
 & \#5 & 23,363 & 25,936 & 30,020 & 33,023 & 35,318 & 35,414 & 39,102 & 40,188 & 42,780 & 45,748 & 49,665 \\
& \textit{mean} & \textbf{23,363} & \textbf{27,819} & \textbf{32,206} & \textbf{36,662} & \textbf{38,377} & \textbf{40,523} & \textbf{42,903} & \textbf{45,277} & \textbf{46,472} & \textbf{49,147} & \textbf{51,413} \\
\midrule
GRACE w/o SA & \#1 & 23,363 & 29,654 & 35,977 & 39,179 & 42,242 & 47,998 & 48,434 & 52,847 & 55,498 & 56,862 & 62,047 \\
 & \#2 & 23,363 & 28,396 & 33,860 & 37,764 & 44,255 & 46,976 & 51,591 & 55,668 & 57,391 & 61,461 & 63,318 \\
 & \#3 & 23,363 & 27,531 & 31,542 & 36,793 & 39,337 & 43,766 & 45,326 & 45,677 & 46,053 & 49,639 & 53,015 \\
 & \#4 & 23,363 & 27,351 & 28,760 & 29,277 & 32,096 & 33,605 & 36,336 & 37,875 & 37,875 & 38,678 & 39,702 \\
 & \#5 & 23,363 & 25,979 & 30,263 & 35,121 & 38,098 & 42,081 & 44,352 & 47,704 & 49,900 & 53,767 & 56,385 \\
& \textit{mean} & \textbf{23,363} & \textbf{27,782} & \textbf{32,080} & \textbf{35,627} & \textbf{39,206} & \textbf{42,885} & \textbf{45,208} & \textbf{47,954} & \textbf{49,343} & \textbf{52,081} & \textbf{54,893} \\
\midrule
HCE & \#1 & 23,363 & 32,915 & 38,273 & 41,752 & 44,381 & 49,548 & 52,610 & 56,740 & 63,844 & 66,625 & 73,615 \\
 & \#2 & 23,363 & 29,322 & 38,187 & 42,009 & 53,734 & 60,065 & 63,802 & 65,870 & 75,473 & 79,360 & 84,941 \\
 & \#3 & 23,363 & 30,322 & 41,693 & 45,093 & 50,864 & 60,111 & 65,989 & 79,524 & 85,341 & 89,506 & 100,125 \\
 & \#4 & 23,363 & 32,426 & 39,852 & 41,776 & 46,321 & 49,347 & 50,283 & 52,977 & 58,889 & 62,042 & 65,535 \\
 & \#5 & 23,363 & 27,628 & 34,764 & 39,003 & 42,735 & 58,545 & 65,660 & 75,336 & 85,567 & 93,475 & 88,748 \\
& \textit{mean} & \textbf{23,363} & \textbf{30,523} & \textbf{38,554} & \textbf{41,927} & \textbf{47,607} & \textbf{55,523} & \textbf{59,669} & \textbf{66,089} & \textbf{73,823} & \textbf{78,202} & \textbf{82,593} \\
\bottomrule
\end{tabular}}

\caption{\textbf{Deployed instruction size in characters at every checkpoint, per replication.} The means at $\ell_0,\ell_6,\ell_8,\ell_{10}$ reproduce Table~\ref{tab:policy-size} up to rounding; the full trajectories underlie the growth-rate estimates in Section~\ref{sec:analysis}.}
\label{tab:raw-size}
\end{table}

\section{Extended Limitations}\label{app:limitations}

This appendix expands the three boundaries summarized in Section~\ref{sec:limitations}.

\paragraph{Single domain, model, and harness.}
All results come from the telecom domain of $\tau^2$-bench with Gemini 2.5 Flash as the agent and GPT-4.1 as the user simulator. Telecom is rule- and procedure-heavy, which is plausibly favorable to a typed-graph substrate; domains whose instructions are less constraint-like, such as open-ended coding or research assistance, may benefit less. We chose telecom because it is the dual-control domain of the benchmark, so both agent and simulated user act on shared state and instruction quality has direct behavioral consequences; the choice does not license claims about other domains, models, or harnesses. Coverage of the three-type schema has likewise been examined only on this instruction (Section~\ref{sec:graph-repr}); instructions dominated by examples, output templates, or tool schemas may need a custom schema before scoped verification is meaningful.

\paragraph{Unmatched internal compute.}
GRACE issues roughly six internal calls per evolution step against one for HCE (Table~\ref{tab:evolution-cost}). The experiments therefore establish that the graph-regularized operator, including its verification calls, sustains improvement where the flat-text operator does not; they do not establish how much of the gap would survive if HCE were given the same call budget, for instance through repeated self-refinement passes. The ablation partially addresses this, since GRACE without SA spends half of GRACE's budget and still collapses, which argues against a pure compute explanation, but a budget-matched flat-text control is the missing experiment.

\paragraph{LLM-judged mechanism audits.}
Contradiction and redundancy counts come from a post-hoc audit by frontier LLM judges. Because structural analysis modifies the graph under criteria related to those the audit applies, and because the judges are themselves LLMs that may share training lineage or failure modes with the evolution operator, part of the measured redundancy reduction could reflect alignment with the judge rather than a change a human would recognize. We therefore use these counts only as correlational support for the mechanism story, never for checkpoint selection or the headline metrics, and an audit with an independent judge model and human agreement is a clear next step.

\section{Implementation Details}\label{app:impl}
\paragraph{Reproducibility.}
The method implementation, the reproduction runner, the frozen task split, all prompts, the reproduction configuration, and the per-replication results in Appendix~\ref{app:raw} are available in the public repository at \url{https://github.com/RedMind-Research/GRACE}. This appendix records the settings needed to reproduce the reported runs (Table~\ref{tab:impl-settings}). The public reproduction runner freezes the procedure (task split, benchmark revision, model routes, seed, prompts, method settings, and metric definitions) and verifies the installed core version and benchmark revision before executing; reproducers regenerate trajectories through the provider APIs.

\paragraph{Models and decoding.}
Task agent, diagnosis, evolution, and prompt-to-graph all use \texttt{gemini-2.5-flash} (Vertex AI); the user simulator is \texttt{gpt-4.1-2025-04-14}. Every method-internal call uses temperature 0 and an output limit of 65{,}536 tokens with a JSON response schema; agent episodes use an 8{,}192-token output limit, at most 200 steps, a 300-second timeout, and at most 10 tool errors. The run seed is 1024. Google has announced the retirement of \texttt{gemini-2.5-flash} for October 16, 2026; reproductions after that date will need to configure a replacement model explicitly, and the prompt-to-graph granularity is model dependent. The deterministic $\tau^2$-bench evaluator scores episodes; no judge model is involved in pass@$k$ or pass\textasciicircum{}$k$.

\paragraph{Structural analysis schedule and stopping rule.}
After operation planning and relation maintenance, the candidate graph is first checked against the network schema and repaired deterministically for up to three schema-repair rounds. Structural analysis then runs for at most three rounds. Round $r$ extracts the local subgraph within radius $k = 3r$ hops of the touched nodes (so $k \in \{3, 6, 9\}$), measured on the relation layer with $\mathsf{identity}$-incident edges excluded as described in Section~\ref{sec:pipeline}, sends it to the detection prompt (Appendix~\ref{app:prompt-sa-detect}), and stops early when the round reports no contradiction and no redundancy. When issues are reported, the repair prompt (Appendix~\ref{app:prompt-sa-repair}) proposes \textsf{ModifyNode}, \textsf{Merge}, \textsf{AddEdge}, \textsf{RemoveEdge}, or \textsf{RemoveNode} operations restricted to the listed issues; the assembled graph is schema-validated and deterministically repaired again before the next round. If the third round still reports issues, the run records that the cap was reached and proceeds with the current candidate. Merge keeps the first node as representative and requires both nodes to share an object type. GRACE without SA skips this stage entirely and changes nothing else.

\paragraph{Deterministic grounding.}
After structural analysis, a deterministic pass applies the schema's grounding policy: every $\mathsf{norm}$ node receives a $\mathsf{supports}$ edge from the top $\mathsf{identity}$ node unless that edge already exists, and any further $\mathsf{identity}$ node is linked to the top identity through $\mathsf{refines}$ with a reachability guard, so that repeated application is idempotent and only schema-admissible, acyclic edges are added. The number of edges added is recorded in the change log. This scaffold is what the neighborhood computation excludes (Section~\ref{sec:pipeline}); it exists so that reconstruction can attribute every norm to the agent's mandate, not to shape validation scope.

\paragraph{Delta reconstruction.}
Reconstruction never regenerates the instruction. The validated node-level changes (added, modified, merged, and removed nodes) are sent with the current instruction to the patch prompt (Appendix~\ref{app:prompt-delta}), which returns \textsf{ReplaceSpan}, \textsf{InsertAfter}, and \textsf{DeleteSpan} operations anchored on verbatim, unique spans of the current text. A deterministic applier executes the operations and copies every untouched character; an operation whose anchor is missing or ambiguous is rejected and logged. The HCE baseline instead returns the complete updated instruction from one call (Appendix~\ref{app:prompt-hce}).

\paragraph{Diagnosis.}
The two-stage diagnosis (Appendices~\ref{app:prompt-diag-1} and~\ref{app:prompt-diag-2}) is shared verbatim by all conditions. Stage one reflects on each failed experience episode in canonical task order and reports system-prompt gaps with a proposed instruction; stage two consolidates the findings into themes with frequency, affected subtask types, root cause, and recommended actions. A batch with no eligible failure produces a \emph{no-update} diagnosis, in which case every condition publishes an unchanged checkpoint without an evolution call. Diagnosis is failure-driven, so its cost varies by condition (\$3.0--\$5.6 per run) even though the procedure is fixed.

\begin{table}[H]
\centering
\footnotesize
\begin{tabular}{ll}
\toprule
\textbf{Setting} & \textbf{Value} \\
\midrule
Benchmark revision & $\tau^2$-bench commit \texttt{c5b2d228} (telecom domain) \\
Experience batches / tasks per batch & 10 / 42 (phase schedule A,A,B,B,A,A,B,B,A,A) \\
Held-out evaluation & 66 tasks $\times$ 3 trials (198 episodes per checkpoint) \\
Method-call decoding & temperature 0, 65{,}536 output tokens, JSON schema \\
P2G fidelity rounds (max) & 10 \\
Schema repair rounds (max) & 3 \\
Structural analysis rounds (max) / radius step & 3 / 3 hops ($k\in\{3,6,9\}$) \\
Initial graph & 335 nodes, 442 edges (identical across temperature-0 constructions) \\
Seed & 1024 \\
\bottomrule
\end{tabular}
\caption{\textbf{Reproduction settings.} Values are read from the frozen reproduction configuration in the public code release.}
\label{tab:impl-settings}
\end{table}

% Generated by tools/gen_prompts_appendix.py from the GRACE repository source. Do not edit by hand.
\section{Prompt Templates}\label{app:prompts}
All prompts below are rendered verbatim from the public reproduction code with the default network schema. Angle-bracket placeholders mark where the harness inserts the graph, subgraph, instruction, or report at run time. Every method-internal call uses temperature 0 and a JSON response contract; the harness rejects non-JSON output and retries within the attempt ledger.

\subsection{Prompt-to-graph decomposition (P2G)}\label{app:prompt-p2g}
\begin{scriptsize}\begin{verbatim}
SYSTEM:
<task>
Build a typed graph G=(V,E) that completely transcribes the supplied system-level
instruction. Decompose the instruction into atomic units. Assign each unit the object
type whose configured definition it satisfies. Add a directed relation only when the
contents satisfy the configured relation definition and type signature. Preserve all
instruction content: omit nothing and invent nothing.
</task>

<network_schema>
<object_types A>
- identity: Identity constitutes the agent as the subject of action: its content fixes the
agent's standing and the mandate under which it acts, establishing the source to which its
conduct is attributed.
- norm: A norm governs the agent's conduct: its content is a standard the agent's action
must conform to, so that any action either satisfies or breaches it.
- knowledge: Knowledge represents the world the agent acts upon: its content describes a
self-contained piece of the task domain that the agent reasons from.
</object_types>

<relation_types R and type signatures>
Notation: each pair (a, b) is a directed edge from type a to type b.
- supports: A relation of grounding. The target content rests on the source content as its
basis, and the edge runs from the grounding content to the content it grounds. Admissible
pairs: (identity, norm), (knowledge, norm), (knowledge, knowledge), (norm, norm).
- refines: A relation of specialization between contents of the same kind. The source is the
target under a narrower scope, and the edge runs from the more specific content to the more
general content. Admissible pairs: (identity, identity), (norm, norm), (knowledge,
knowledge). Constraints: acyclic.
- sequence: A relation of procedural order over norms. The source content is carried out
before the target content in execution. Admissible pairs: (norm, norm). Constraints:
acyclic.
</relation_types>
</network_schema>

<atomicity>
Each node is one independently checkable instruction unit. Split multi-step procedures,
exceptions, conditional branches, and enumerations into atomic units; do not fragment one
atomic unit. Use relations to retain grounding, specialization, and order represented by
the source instruction.
</atomicity>

<output_format>
Return valid JSON only:
{"nodes":[{"id":"N001","type":"identity|norm|knowledge","content":"..."}],
 "edges":[{"source":"N001","target":"N002","relation":"supports|refines|sequence"}]}
Do not include commentary or any other top-level field.
</output_format>

USER:
<system_instruction>
<system-level instruction>
</system_instruction>

Decompose this instruction into the configured typed graph. Output JSON only.
\end{verbatim}\end{scriptsize}

\subsection{P2G fidelity analysis}\label{app:prompt-p2g-fidelity}
\begin{scriptsize}\begin{verbatim}
SYSTEM:
<task>
Compare an original system-level instruction with a typed graph claimed to transcribe it.
Report concrete mismatches only: missing instruction units, content that adds/removes/
distorts meaning, incorrect object typing, incorrect represented relations, or clearly
implied configured relations that are absent. Do not flag equivalent wording or optional
relations not clearly implied by the source.
</task>

<network_schema>
<object_types A>
- identity: Identity constitutes the agent as the subject of action: its content fixes the
agent's standing and the mandate under which it acts, establishing the source to which its
conduct is attributed.
- norm: A norm governs the agent's conduct: its content is a standard the agent's action
must conform to, so that any action either satisfies or breaches it.
- knowledge: Knowledge represents the world the agent acts upon: its content describes a
self-contained piece of the task domain that the agent reasons from.
</object_types>

<relation_types R and type signatures>
Notation: each pair (a, b) is a directed edge from type a to type b.
- supports: A relation of grounding. The target content rests on the source content as its
basis, and the edge runs from the grounding content to the content it grounds. Admissible
pairs: (identity, norm), (knowledge, norm), (knowledge, knowledge), (norm, norm).
- refines: A relation of specialization between contents of the same kind. The source is the
target under a narrower scope, and the edge runs from the more specific content to the more
general content. Admissible pairs: (identity, identity), (norm, norm), (knowledge,
knowledge). Constraints: acyclic.
- sequence: A relation of procedural order over norms. The source content is carried out
before the target content in execution. Admissible pairs: (norm, norm). Constraints:
acyclic.
</relation_types>
</network_schema>

<output_format>
Return valid JSON only with exactly these lists:
{"missing_node":["source unit"],
 "unfaithful":[{"node_id":"N001","issue":"..."}],
 "mistyped":[{"node_id":"N001","should_be":"identity|norm|knowledge","reason":"..."}],
 "wrong_relation":[{"source":"N001","target":"N002","should_be":"supports|refines|sequence",
"reason":"..."}],
 "missing_relation":[{"source":"N001","target":"N002","relation":"supports|refines|sequence"
,"reason":"..."}]}
Use empty lists when no issue exists.
</output_format>

USER:
<system_instruction>
<system-level instruction>
</system_instruction>

<graph>
"<graph JSON>"
</graph>

Judge fidelity and completeness. Output JSON only.
\end{verbatim}\end{scriptsize}

\subsection{P2G fidelity repair}\label{app:prompt-p2g-repair}
\begin{scriptsize}\begin{verbatim}
SYSTEM:
<task>
Repair the current typed graph so it faithfully and completely transcribes the original
instruction. Address only the listed schema/fidelity issues with minimal targeted edits;
leave correct graph content unchanged. Never repeat an operation already rejected by the
deterministic assembler.
</task>

<network_schema>
<object_types A>
- identity: Identity constitutes the agent as the subject of action: its content fixes the
agent's standing and the mandate under which it acts, establishing the source to which its
conduct is attributed.
- norm: A norm governs the agent's conduct: its content is a standard the agent's action
must conform to, so that any action either satisfies or breaches it.
- knowledge: Knowledge represents the world the agent acts upon: its content describes a
self-contained piece of the task domain that the agent reasons from.
</object_types>

<relation_types R and type signatures>
Notation: each pair (a, b) is a directed edge from type a to type b.
- supports: A relation of grounding. The target content rests on the source content as its
basis, and the edge runs from the grounding content to the content it grounds. Admissible
pairs: (identity, norm), (knowledge, norm), (knowledge, knowledge), (norm, norm).
- refines: A relation of specialization between contents of the same kind. The source is the
target under a narrower scope, and the edge runs from the more specific content to the more
general content. Admissible pairs: (identity, identity), (norm, norm), (knowledge,
knowledge). Constraints: acyclic.
- sequence: A relation of procedural order over norms. The source content is carried out
before the target content in execution. Admissible pairs: (norm, norm). Constraints:
acyclic.
</relation_types>
</network_schema>

<editing_algebra>
Available operations are AddNode, ModifyNode, RemoveNode, AddEdge, and RemoveEdge.
ModifyNode changes content only. Correct a mistyped node by replacing it. Every operation
must satisfy the configured schema; if no faithful admissible fix exists, leave that issue
unresolved rather than inventing content.
</editing_algebra>

<output_format>
Return valid JSON only as {"operations":[...]}. Each operation uses the public GRACE
operation fields and may include an audit-only "fixes" field.
</output_format>

USER:
<system_instruction>
<system-level instruction>
</system_instruction>
<graph>"<graph JSON>"</graph>
<issues>"<issues JSON>"</issues>
<rejected_operations>"<rejected operations JSON>"</rejected_operations>

Propose minimal admissible repairs. Output JSON only.
\end{verbatim}\end{scriptsize}

\subsection{Operation planning}\label{app:prompt-op-plan}
\begin{scriptsize}\begin{verbatim}
SYSTEM:
<task>
Translate a diagnosis report into minimal graph-native policy edits. First locate graph
content that already addresses each finding. Propose an edit only for a genuine omission,
misstatement, or removal. Ground every operation in a diagnosis finding. This stage edits
nodes; consolidation through Merge is reserved for structural repair.
</task>
<network_schema><object_types A>
- identity: Identity constitutes the agent as the subject of action: its content fixes the
agent's standing and the mandate under which it acts, establishing the source to which its
conduct is attributed.
- norm: A norm governs the agent's conduct: its content is a standard the agent's action
must conform to, so that any action either satisfies or breaches it.
- knowledge: Knowledge represents the world the agent acts upon: its content describes a
self-contained piece of the task domain that the agent reasons from.
</object_types>

<relation_types R and type signatures>
Notation: each pair (a, b) is a directed edge from type a to type b.
- supports: A relation of grounding. The target content rests on the source content as its
basis, and the edge runs from the grounding content to the content it grounds. Admissible
pairs: (identity, norm), (knowledge, norm), (knowledge, knowledge), (norm, norm).
- refines: A relation of specialization between contents of the same kind. The source is the
target under a narrower scope, and the edge runs from the more specific content to the more
general content. Admissible pairs: (identity, identity), (norm, norm), (knowledge,
knowledge). Constraints: acyclic.
- sequence: A relation of procedural order over norms. The source content is carried out
before the target content in execution. Admissible pairs: (norm, norm). Constraints:
acyclic.
</relation_types></network_schema>
<scope>
Allowed operations: AddNode, ModifyNode, RemoveNode. AddNode may attach configured edges.
Do not propose Merge, AddEdge, or RemoveEdge as a top-level operation here.
</scope>
<operation_contracts>
AddNode: {"op":"AddNode","type":"...","content":"...","attach_edges":[{"source":"NEW","targe
t":"existing_id","relation":"..."}]}
ModifyNode: {"op":"ModifyNode","node_id":"...","new_content":"..."}
RemoveNode: {"op":"RemoveNode","node_id":"..."}
</operation_contracts>
<output_format>
Return JSON only as {"operations":[...]}. Every item must use the exact discriminator
field `op` (never `operation`), for example {"op":"ModifyNode",...}.
</output_format>

USER:
<graph>"<graph JSON>"</graph>
<diagnosis_report><diagnosis report></diagnosis_report>
Propose minimal node-level edits. Output JSON only.
\end{verbatim}\end{scriptsize}

\subsection{Relation maintenance}\label{app:prompt-rel-maint}
\begin{scriptsize}\begin{verbatim}
SYSTEM:
<task>
Review relations incident to the changed nodes after node editing. Add a relation only when
the endpoint contents satisfy its configured meaning/signature; remove only a relation that
no longer holds. Preserve unrelated relations.
</task>
<network_schema><object_types A>
- identity: Identity constitutes the agent as the subject of action: its content fixes the
agent's standing and the mandate under which it acts, establishing the source to which its
conduct is attributed.
- norm: A norm governs the agent's conduct: its content is a standard the agent's action
must conform to, so that any action either satisfies or breaches it.
- knowledge: Knowledge represents the world the agent acts upon: its content describes a
self-contained piece of the task domain that the agent reasons from.
</object_types>

<relation_types R and type signatures>
Notation: each pair (a, b) is a directed edge from type a to type b.
- supports: A relation of grounding. The target content rests on the source content as its
basis, and the edge runs from the grounding content to the content it grounds. Admissible
pairs: (identity, norm), (knowledge, norm), (knowledge, knowledge), (norm, norm).
- refines: A relation of specialization between contents of the same kind. The source is the
target under a narrower scope, and the edge runs from the more specific content to the more
general content. Admissible pairs: (identity, identity), (norm, norm), (knowledge,
knowledge). Constraints: acyclic.
- sequence: A relation of procedural order over norms. The source content is carried out
before the target content in execution. Admissible pairs: (norm, norm). Constraints:
acyclic.
</relation_types></network_schema>
<scope>Allowed operations: AddEdge and RemoveEdge only.</scope>
<operation_contracts>
AddEdge: {"op":"AddEdge","source":"...","target":"...","relation":"..."}
RemoveEdge: {"op":"RemoveEdge","source":"...","target":"...","relation":"..."}
</operation_contracts>
<output_format>
Return JSON only as {"operations":[...]}. Every item must use the exact discriminator
field `op` (never `operation`).
</output_format>

USER:
<graph>"<graph JSON>"</graph>
<changed_node_ids>"<changed node ids>"</changed_node_ids>
Propose relation maintenance. Output JSON only.
\end{verbatim}\end{scriptsize}

\subsection{Structural analysis (detection)}\label{app:prompt-sa-detect}
\begin{scriptsize}\begin{verbatim}
SYSTEM:
<task>
Detect every contradiction and redundancy between distinct nodes in the supplied local
typed graph. A contradiction has overlapping scope and no coherent joint reading. A
redundancy requires the same object type, overlapping scope, semantic subsumption, and no
configured structural reason to retain both. Detection only; propose no repair.
</task>
<network_schema><object_types A>
- identity: Identity constitutes the agent as the subject of action: its content fixes the
agent's standing and the mandate under which it acts, establishing the source to which its
conduct is attributed.
- norm: A norm governs the agent's conduct: its content is a standard the agent's action
must conform to, so that any action either satisfies or breaches it.
- knowledge: Knowledge represents the world the agent acts upon: its content describes a
self-contained piece of the task domain that the agent reasons from.
</object_types>

<relation_types R and type signatures>
Notation: each pair (a, b) is a directed edge from type a to type b.
- supports: A relation of grounding. The target content rests on the source content as its
basis, and the edge runs from the grounding content to the content it grounds. Admissible
pairs: (identity, norm), (knowledge, norm), (knowledge, knowledge), (norm, norm).
- refines: A relation of specialization between contents of the same kind. The source is the
target under a narrower scope, and the edge runs from the more specific content to the more
general content. Admissible pairs: (identity, identity), (norm, norm), (knowledge,
knowledge). Constraints: acyclic.
- sequence: A relation of procedural order over norms. The source content is carried out
before the target content in execution. Admissible pairs: (norm, norm). Constraints:
acyclic.
</relation_types></network_schema>
<output_format>
Return JSON only:
{"contradictions":[{"u_id":"...","v_id":"...","reason":"..."}],
 "redundancies":[{"u_id":"...","v_id":"...","reason":"..."}]}
</output_format>

USER:
<local_graph>"<local subgraph JSON>"</local_graph>
Detect local contradictions and redundancies. Output JSON only.
\end{verbatim}\end{scriptsize}

\subsection{Structural repair}\label{app:prompt-sa-repair}
\begin{scriptsize}\begin{verbatim}
SYSTEM:
<task>
Resolve each listed local contradiction or redundancy with minimal schema-admissible graph
change. Use ModifyNode/remove operations to restore coherence and Merge or RemoveNode to
consolidate true redundancy. Each operation must reference a listed issue.
</task>
<network_schema><object_types A>
- identity: Identity constitutes the agent as the subject of action: its content fixes the
agent's standing and the mandate under which it acts, establishing the source to which its
conduct is attributed.
- norm: A norm governs the agent's conduct: its content is a standard the agent's action
must conform to, so that any action either satisfies or breaches it.
- knowledge: Knowledge represents the world the agent acts upon: its content describes a
self-contained piece of the task domain that the agent reasons from.
</object_types>

<relation_types R and type signatures>
Notation: each pair (a, b) is a directed edge from type a to type b.
- supports: A relation of grounding. The target content rests on the source content as its
basis, and the edge runs from the grounding content to the content it grounds. Admissible
pairs: (identity, norm), (knowledge, norm), (knowledge, knowledge), (norm, norm).
- refines: A relation of specialization between contents of the same kind. The source is the
target under a narrower scope, and the edge runs from the more specific content to the more
general content. Admissible pairs: (identity, identity), (norm, norm), (knowledge,
knowledge). Constraints: acyclic.
- sequence: A relation of procedural order over norms. The source content is carried out
before the target content in execution. Admissible pairs: (norm, norm). Constraints:
acyclic.
</relation_types></network_schema>
<scope>
Allowed operations: ModifyNode, Merge, AddEdge, RemoveEdge, RemoveNode. Merge requires the
same object type and keeps u_id as representative.
</scope>
<operation_contracts>
ModifyNode: {"op":"ModifyNode","node_id":"...","new_content":"..."}
Merge: {"op":"Merge","u_id":"...","v_id":"...","new_content":"..."}
AddEdge/RemoveEdge: {"op":"AddEdge","source":"...","target":"...","relation":"..."}
RemoveNode: {"op":"RemoveNode","node_id":"..."}
</operation_contracts>
<output_format>
Return JSON only as {"operations":[...]}. Every item must use the exact discriminator
field `op` (never `operation`).
</output_format>

USER:
<local_graph>"<local subgraph JSON>"</local_graph>
<issues>"<issues JSON>"</issues>
Propose minimal repairs. Output JSON only.
\end{verbatim}\end{scriptsize}

\subsection{Delta reconstruction (patch operations)}\label{app:prompt-delta}
\begin{scriptsize}\begin{verbatim}
SYSTEM:
<task>
Emit localized operations that update a system instruction for the supplied validated
node-level changes. A deterministic applier copies every untouched character. Do not
reproduce or rewrite unaffected text.
</task>
<operations>
ReplaceSpan: {"op":"ReplaceSpan","anchor":"...","new_text":"..."}
InsertAfter: {"op":"InsertAfter","anchor":"...","new_text":"..."}
DeleteSpan: {"op":"DeleteSpan","anchor":"..."}
</operations>
<anchor_rules>
Every anchor is copied verbatim from the current instruction and appears exactly once.
Use the shortest unique real span. For an addition, anchor at the closest related existing
text. Never invent a heading or use paraphrased graph content as an anchor.
</anchor_rules>
<output_format>
Return valid JSON only as {"operations":[...]}. Every item must use the exact
discriminator field "op" (never "operation").
</output_format>

USER:
<system_instruction><current instruction text></system_instruction>
<node_changes>"<validated node changes JSON>"</node_changes>
Produce minimal patch operations. Output JSON only.
\end{verbatim}\end{scriptsize}

\subsection{HCE baseline: single-call holistic update}\label{app:prompt-hce}
\begin{scriptsize}\begin{verbatim}
SYSTEM:
<task>
Given the system prompt and a diagnosis report, produce the updated system prompt in one LLM
call. Apply the diagnosis as incremental updates to the system prompt. This is not a
rewrite. Make targeted additions and refinements that address diagnosed failure patterns
while keeping unrelated content intact. Return only the complete updated system prompt.
</task>

<input>
1. system_prompt: the deployed system prompt used during the experience batch.
2. diagnosis_report: a markdown report consolidating failure themes from trajectories. Each
theme includes a title, frequency, affected subtask types, root cause, and recommended
actions.
</input>

<procedure>
This is one LLM call. Carry out the three phases below in order as a single continuous
process that produces one final output. Do not pause for external input, do not emit
intermediate artifacts, and produce nothing until the Verifier phase has completed.
1. Planner: identify the targeted additions and refinements needed to address the diagnosis.
2. Executor: apply those additions and refinements as incremental updates to the system
prompt.
3. Verifier: check the drafted system prompt for conflicts and same-level duplication, then
revise only to resolve those issues.
</procedure>

<Planner>
Use the diagnosis report to identify what needs improvement and why. For every planned
update, connect the diagnosed failure pattern to the system prompt gap it addresses. Plan
additions when the system prompt lacks guidance for a diagnosed gap. Plan refinements when
existing guidance is incomplete, imprecise, ambiguous, mis-scoped, or poorly placed for the
diagnosed case.
</Planner>

<Executor>
Draft the updated system prompt by applying the planned additions and refinements as
incremental updates to the original system prompt.

Add new guidance near related existing content. Revise nearby existing wording when the
planned update concerns an existing instruction. Keep unrelated content intact.

Preserve section order, headings, list formatting, inline conventions, XML tags, and the
ordering of existing sections unless a planned update directly concerns that local text.
</Executor>

<Verifier>
Examine the drafted system prompt for conflicts and same-level duplication introduced or
exposed by the update.

A conflict is a pair of statements that apply to the same situation and cannot both be
followed.

Same-level duplication is a pair of statements that gives the same guidance at the same
level of specificity.

Revise the draft only to resolve conflicts or same-level duplication. Do not reorganize
unrelated content, compress the prompt, or turn the update into a broader rewrite.
</Verifier>

<output_format>
Return the complete updated system prompt as a single plain-text string, with no diff,
summary, commentary, or markdown fencing.
</output_format>

USER:
<system_prompt>
{current_prompt}
</system_prompt>

<diagnosis_report>
{diagnosis_report}
</diagnosis_report>

Update this system prompt based on the diagnosis. Output the complete updated system prompt.
\end{verbatim}\end{scriptsize}

\subsection{Diagnosis stage 1: per-failure reflection}\label{app:prompt-diag-1}
\begin{scriptsize}\begin{verbatim}
SYSTEM:
<task>
You are given a failed agent trajectory and the system prompt the agent operated under.
Produce a comprehensive per-failure diagnosis. Determine how the agent handled each required
subtask, identify every system-prompt gap that contributed to the failure, and report each
gap as one finding with a concrete instruction the system prompt should add or change. Work
through the full trajectory and required subtasks using the analysis approach below, follow
the rules, and return the output in the format below.
</task>

<input>
1. system_prompt: the system prompt the agent operated under.
2. trajectory: the complete trajectory, including the agent's actions, tool calls,
observations, and user turns.
3. subtasks: the required subtasks for this task.
</input>

<analysis_approach>
- Trace the full trajectory in order, including user turns, agent actions, tool calls,
observations, user-facing instructions, tool errors, and transfers.
- For each required subtask, decide whether the agent attempted it, completed it, left it
unresolved, or left it unaddressed before failure.
- Mark a subtask as attempted when the trajectory contains an explicit action, tool call,
user-facing instruction, or diagnostic step targeting that subtask's required outcome.
- For each unresolved or unaddressed subtask, identify what system-prompt guidance,
procedure, knowledge, decision criterion, or capability description is missing or
inadequate.
- Also identify any issue that affects multiple subtasks or the overall failure trajectory.
- Report all distinct gaps that contributed to the failure.
</analysis_approach>

<global_rules>
- Ground every finding in behavior observed in the trajectory.
- Cover every distinct system-prompt gap the trajectory reveals, through to the last
failure-relevant gap.
- Prefer one finding per distinct missing or inadequate instruction.
- Include multiple affected subtasks in the same finding when one prompt gap explains
several failures.
- Give each finding a concrete proposed instruction that could be added to or used to revise
the system prompt.
- When the trajectory suggests that the system prompt lacks guidance for obtaining needed
state, carrying out an allowed operation, choosing the next step, or verifying an outcome,
describe the missing guidance or capability concretely in the proposed instruction.
</global_rules>

<output_format>
Return valid JSON only:
{
  "task_id": "<task identifier>",
  "subtasks_attempted": ["<required subtask the agent attempted through an action, tool
call, user-facing instruction, or diagnostic step>"],
  "subtasks_missed": ["<required subtask with no meaningful attempt before failure>"],
  "findings": [
    {"finding_id": "<e.g. F1>", "affected_subtasks": ["<subtask affected by this gap>"],
"description": "<2-3 sentences: what went wrong and what guidance, procedure, knowledge,
decision criterion, or capability description is missing or inadequate>",
"proposed_instruction": "<concrete instruction text to add or modify in the system prompt>"}
  ]
}
</output_format>

USER:
<system_prompt>
{system_prompt}
</system_prompt>

<task_id>
{task_id}
</task_id>

<required_subtasks>
{subtasks_list}
</required_subtasks>

<trajectory>
{trajectory}
</trajectory>

Analyze all gaps in this failed trajectory. For each required subtask that was unresolved or
unaddressed, identify what guidance is missing or inadequate in the system prompt. Output
valid JSON only.
\end{verbatim}\end{scriptsize}

\subsection{Diagnosis stage 2: batch synthesis}\label{app:prompt-diag-2}
\begin{scriptsize}\begin{verbatim}
SYSTEM:
<task>
You are given the per-failure analyses for one batch of failed trajectories. Produce a
domain-general batch diagnosis for downstream prompt evolution. Consolidate the findings
into themes, where each theme represents one underlying system-prompt gap and one coherent
fix target. Follow the synthesis approach below, follow the rules, and return the output in
the format below.
</task>

<input>
reflections: a JSON list of per-failure analyses, each containing findings that identify
system-prompt gaps.
</input>

<synthesis_approach>
- Start from the input findings, and preserve every finding in the final diagnosis.
- Group findings into the same theme only when they point to the same missing or inadequate
system-prompt instruction.
- Split findings into separate themes when they require different prompt instructions, even
if they affect the same subtask or tool.
- Merge findings when their proposed instructions are equivalent, overlapping, or one
clearly subsumes the other.
- Keep the most relevant proposed instruction from the supporting findings, preserving its
concrete names and wording where possible.
- Keep each theme as specific as the gap it describes, naming the behavior, tool, action, or
condition the system prompt should govern.
</synthesis_approach>

<theme_fields>
- frequency: the count of distinct failed trajectories that show this gap, counting each
trajectory once. It records how widespread the gap is, and every theme remains equally
required to be addressed.
- affected_subtask_types: the subtask types the gap touches, drawn from the findings.
- root_cause: the shared system-prompt gap, grounded in the findings that support the theme.
- recommended_actions: concrete, self-contained instructions that the next evolution step
can add to or use to revise the system prompt.
</theme_fields>

<global_rules>
- Build every theme from findings that describe behavior observed in the trajectories.
- Place every finding in exactly one theme, and keep all findings.
- Represent each distinct system-prompt gap once.
- A theme drawn from a single trajectory is valid and equally required to be addressed.
- Let recommended_actions hold several instructions only when one coordinated fix requires
them.
- Make each recommended action specific enough to be inserted into a system prompt without
additional interpretation.
- When a supporting Phase 1 finding contains concrete tool names, action names, condition
names, policy section names, or other named capabilities, preserve those names exactly as
written in the theme root_cause or recommended_actions.
</global_rules>

<output_format>
Return valid JSON only:
{
  "themes": [
    {"theme_id": "<theme id>", "title": "<concise theme name, max 10 words>", "frequency":
<distinct trajectory count>, "affected_subtask_types": ["<subtask type affected>"],
"root_cause": "<2-3 sentences synthesizing the shared prompt gap and its evidence>",
"recommended_actions": ["<specific instruction describing the guidance the system prompt
should hold>"]}
  ],
  "summary": "<overall diagnosis in 2-3 sentences>"
}
</output_format>

USER:
<reflections>
{num_reflections} per-failure analyses to consolidate:
{reflections_json}
</reflections>

Consolidate these Phase 1 findings into one structured batch diagnosis. Preserve every
finding, cluster by underlying system-prompt gap, and output valid JSON only.
\end{verbatim}\end{scriptsize}

\section{Use of Large Language Models}\label{app:llm-use}
LLMs are central to the research methodology of this work. Gemini 2.5 Flash serves as the agent model and GPT-4.1 as the user simulator (Section~\ref{sec:setup}); LLM calls also perform operation planning, relation maintenance, structural analysis, and delta reconstruction inside the GRACE pipeline (Section~\ref{sec:method}), and frontier LLM judges act as post-hoc auditors (Appendix~\ref{app:audit}). Beyond the methodology, an LLM assistant was used during manuscript preparation to polish author-drafted text, to help write and check analysis scripts, and to cross-check reported numbers against the underlying result tables. The research questions, method, experimental design, and all conclusions are the authors', and every LLM-assisted passage was reviewed and revised by the authors, who take full responsibility for the content.

\end{document}